\documentclass{article}
\usepackage{graphicx,xcolor,amssymb,amsmath} 

\usepackage[totalwidth=15.75cm,totalheight=22.275cm]{geometry}
\usepackage{hyperref}
\usepackage{url}
\usepackage{bbm}
\usepackage{pgffor} 
\usepackage{booktabs}
\usepackage{mathrsfs}
\usepackage{authblk}
\usepackage{mleftright}
\usepackage{siunitx}
\usepackage[normalem]{ulem}

\usepackage{xr-hyper}
\usepackage{hyperref}

\usepackage[nospace,sort,compress]{cite}

\newcommand{\R}{\ensuremath{\mathbb{R}}}

\newcommand{\pow}{\operatorname{pow}}

\DeclareMathOperator{\argmin}{\arg\!\min}
\DeclareMathOperator{\softmax}{\mathrm{softmax}}

\newcommand{\GT}{\ensuremath{\mathrm{GT}}}

\newcommand{\fa}{\mathfrak{M}}
\newcommand{\fx}{\mathfrak{Y}}
\newcommand{\apdmatrix}{M}
\newcommand{\apdweight}{w}

\title{
 Fitting Generalized Power Diagrams to 3D Image Data: \\ A Prerequisite for Virtual Materials Testing
}

\author[1]{Andreas~Alpers\thanks{Corresponding author: andreas.alpers@liverpool.ac.uk}}
\author[2,3]{Orkun~Furat} 
\author[4]{Christian~Jung} 
\author[2,5]{Matthias~Neumann}
\author[4]{Claudia~Redenbach} 
\author[1]{Aigerim~Saken}
\author[2]{Volker~Schmidt}

\affil[1]{Department of Mathematical Sciences, University of Liverpool, Liverpool L69 7ZL, UK}
\affil[2]{Institute of Stochastics, Ulm University, 89069 Ulm, Germany}
\affil[3]{SDU Applied AI and Data Science Unit, University of Southern Denmark, 5230 Odense, Denmark}
\affil[4]{Department of Mathematics, RPTU University Kaiserslautern-Landau, \newline 67663 Kaiserslautern, Germany}
\affil[5]{Institute of Statistics, Graz University of Technology, 8010 Graz, Austria}
\date{\today}

\begin{document}

\maketitle

\begin{abstract} 
\noindent
This paper reviews algorithmic and modeling approaches for fitting generalized power diagrams to three-dimensional image data, a key step in virtual materials testing (VMT). Beyond their practical relevance to materials science, these tessellation models connect to several active areas of applied mathematics, including optimization, computational geometry, stochastic modeling, and optimal transport. Their formulation combines concepts from convex analysis and geometric clustering, offering a rich interplay between theory and computation. We survey recent applications and quantitatively compare algorithmic strategies for fitting Voronoi diagrams, power diagrams, and generalized balanced power diagrams (GBPDs), including linear and nonlinear programming, stochastic optimization via the cross-entropy method, and gradient-based approaches. Comparative results on real datasets illustrate trade-offs between algorithmic complexity and model accuracy.
\end{abstract}



\newpage

\section{Introduction}\label{sec:introduction}



The past two decades have witnessed the emergence and rapid growth of 3D materials science, driven by progress in both experimental instrumentation and computational modeling; see, e.g.,~\cite{materialsscience, jensen2012perspectives, poulsenbook,ctbook}. An area of focus in this field is the quantitative understanding of the influence of material microstructure on the materials macroscopic (functional) properties. With recent advances in experimental and computational methods, these so-called structure-property relationships are increasingly pursued through \emph{virtual materials testing} (VMT)~\cite{OKEREKE2014637}.
This approach integrates imaging techniques, geometric and stochastic 3D modeling, and numerical simulation to establish structure–property relationships in a systematic and resource-efficient manner, since VMT can reduce experimental efforts to derive these relationships.
Within the workflow of VMT, experimentally acquired microscopic or tomographic images provide the database for the calibration of mathematical models (e.g., stochastic, geometric, and numerical models) that can generate virtual, but realistic 3D images of material microstructures~\cite{schladitz-particles,lautensack2008random,furat.2021, jeulin.2021}. In addition, parameters of calibrated stochastic 3D models can be varied to generate even further 3D images that exhibit structural scenarios that have not yet been observed. The generated 3D image data can serve as input for numerical simulations of macroscopic material properties that complement or even replace costly and time-consuming experimental tests. By means of regression analysis, the structural descriptors of the generated 3D images can be related to simulated macroscopic properties to derive structure-property relationships, see~\cite{prifling2021} and the review in Chapter~5 of~\cite{holzer.2023}. The latter, on the other hand, can be deployed for design purposes, i.e., to identify microstructures with desirable properties \cite{rassloff2025inverse}.

Since materials can possess microstructures belonging to distinct geometric families, the mathematical models within the VMT workflow must be tailored to the corresponding microstructural class. Typical examples include cellular, granular, fibrous, and polycrystalline materials, whose differing topologies and spatial correlations require specific geometric or stochastic modeling approaches. Many technical and natural materials---such as metals, alloys, ceramics, and rocks---are \emph{polycrystalline}, meaning that they consist of a large number of crystalline grains—regions of uniform lattice orientation—separated by grain boundaries.
For these types of materials, VMT can be deployed to investigate, for example, phase transformations, plasticity, and grain growth. The developments in computational methods have enabled unprecedented insights into the microstructural evolution of polycrystalline materials (see, e.g.,~\cite{science1, science0, science2, science3, Zhang.2020}) and have catalyzed the creation of modeling frameworks that balance geometric fidelity and computational cost (see, e.g.,~\cite{kuhn2008modeling, lyckegaard.2011, alpers.2015, mckenna2014grain, groeber2014dream}).
However, deploying the VMT workflow directly on experimentally measured 3D image data can be impractical due to the high data dimensionality. In particular, such 3D images may contain billions of voxels (the volumetric analogs of pixels)~\cite{ctbook,ohser.2009} which makes direct stochastic 3D modeling difficult. Therefore, an essential step within the VMT workflow is the representation of measured grain structures with parametric geometry models, which can significantly reduce the complexity for subsequent stochastic 3D modeling tasks~\cite{alpers.2023b,petrich.2021,neper-paper,spettl.2016}.

For polycrystalline materials, space-filling partitions of the volume of interest, i.e., \emph{tessellations}, are often an excellent choice for efficiently representing 3D image data. 
In particular, tessellation-based representations provide a sparse but accurate geometric description of polycrystalline microstructures observed in image data, enabling efficient storage of grain boundary geometries and fast computation of geometric and topological quantities in subsequent simulations. 
 Similarly to polycrystalline materials, foams also form a space-filling system of almost polyhedral sets. Therefore, the 3D microstructure of both material classes can be described with similar mathematical models.
Moreover, such fitted models can suppress high-frequency noise in experimental data, facilitating more robust geometrical and statistical analyzes. Thus, fitting tessellations to 3D image data facilitates stochastic 3D modeling and consequently VMT. In particular, mathematical models fitted to the image data then allow a systematic investigation of properties such as elasticity, permeability, or heat conduction \cite{Jung2022,Foehst,REDENBACH201270}.

Among tessellation models, \emph{power diagrams}—also known as \emph{Laguerre} or \emph{Voronoi–Laguerre} diagrams~\cite{imai1985voronoi,overview3}—represent the most widely used classes for modeling materials microstructures.  The unweighted special case, the \emph{Voronoi diagram}, has a long history of use in this context, tracing back—conceptually at least—to Johannes Kepler’s 1611 essay \emph{De nive sexangula} (\emph{The Six-Cornered Snowflake}; see~\cite{kepler}), where Kepler discussed the efficiency of space-filling honeycomb structures (the planar Voronoi tessellation of a triangular lattice). For historical accounts, see~\cite{liebling2012voronoi} and the monographs~\cite{aurenhammerbook,okabebook}.
Furthermore, power diagrams have been used, e.g., in grain growth models~\cite{graingrowth1,riosglicksman07,telley96a,telley96b} and in mechanical simulations where the local neighborhood structure of grains is of primary importance~\cite{mikadawson98}. Heuristics have been developed in which grains are characterized by seed points and associated volumes~\cite{kuhn2008modeling,lyckegaard.2011,laguerre2,telley92,altendorf.2014} (see also~\cite{inverting} for the inverse problem).

Because power diagrams are made up of convex sets, more general tessellation models have been introduced to overcome this constraint. 
In particular, the use of local metrics—introducing spatially varying distance measures in the Voronoi context—has been studied in~\cite{mesh2,bwy-08}. 
During the past decade, a further generalization, \emph{generalized balanced power diagrams} (GBPDs; also known as \emph{anisotropic power diagrams}), has been proposed independently as a flexible framework for modeling microstructures, starting, to the best of our knowledge, 
with~\cite{alpers.2015} and~\cite{altendorf.2014}. Several subsequent studies, including~\cite{sedivy,ulm2,spettl.2016}, have shown that GBPDs achieve an accuracy in reproducing experimentally observed microstructures that surpasses previously considered tessellation models, a conclusion further substantiated by the results presented here and discussed in later sections of the present paper. However, before turning to algorithmic and modeling details, it is instructive to highlight why the study of such tessellation models is of particular interest from an applied mathematics perspective. Beyond their relevance to materials science, power diagram–based models connect to several active mathematical areas, including optimization, computational geometry, stochastic modeling, mathematical imaging, and the analysis of high-dimensional data. Their formulation naturally involves concepts from  convex analysis, geometrical clustering, and optimal transport, providing a rich interplay between theory and computation.

To illustrate the breadth of this framework, we briefly review applications of power diagrams and their generalizations in areas such as materials testing, imaging (superpixel generation), stochastic geometry, stereology, and others, before returning to their algorithmic realization—primarily for virtual materials testing.
We investigate several algorithmic approaches for fitting tessellation models, namely Voronoi diagrams, Laguerre diagrams, and GBPDs to volumetric image data of materials such as polycrystals and foams.  We explore a range of optimization techniques, from linear programming methods to stochastic and gradient-based algorithms, evaluating their effectiveness in capturing complex microstructural geometries. Through comparative experiments on real datasets, we highlight the trade-offs between the complexity of the optimization methods and the accuracy of the resulting tessellation approximations, aiming to guide the selection of suitable approaches for various application scenarios. 

The remainder of this paper is organized as follows. 
Section~\ref{sec.tes.mod} introduces the mathematical definition of power diagrams and their generalizations.
Section~\ref{sect:appl} reviews their applications in different scientific domains. Then, Section~\ref{sect:alg} describes the optimization-based fitting algorithms considered in this paper. Section~\ref{sect:data} presents the datasets used.  Section~\ref{sec:comparison} details the performance evaluation measures and reports the comparative fitting results, analyzing the observed trade-offs between reconstruction accuracy and computational complexity. Finally, Section~\ref{sect:summary} summarizes the main findings.

\vspace{-0.4cm}

\section{Basic notions and notation}\label{sec.tes.mod}

Tessellations can generally be defined in Euclidean spaces $\R^d$ for an arbitrary dimension $d\geq1$. However, in the present paper, we restrict our attention to the case $d=3$. 
Let $\phi\subseteq \mathbb{R}^3$ be a non-empty locally finite set of sites in $\R^3$, where locally finite means that any bounded subset of $\mathbb{R}^3$ only contains a finite number of elements of~$\phi.$ The points $x\in\phi$ are called generators. For our application, it is sufficient to consider finite sets of generator points such that the sites are contained in a bounded observation window $V \subset \mathbb{R}^3$. 
The {\it Voronoi tessellation} of $\phi$, also said to be a {\it Voronoi diagram}, is the collection of all sets of the form
\begin{equation}\label{eq:Voronoi}
    C(x,\phi) = \{ y \in \R^3 \, :\, ||y-x|| \le ||y-x'|| \quad \text{for all } x' \in \phi\}, 
\end{equation} 
for $x\in\phi$, with $||\cdot||$ denoting the Euclidean norm in $\R^3$. 
Generalizations of the Voronoi tessellation are obtained by assigning real-valued weights $w$ to the generators $x$ which are then incorporated into the distance measure. Let now $\phi\subseteq \mathbb{R}^3 \times \R$ be a locally finite set of weighted generators. 
Then, the {\it Laguerre tessellation} of $\phi$ is the collection of all non-empty sets of the form
\begin{equation}\label{eq:Laguerre}
    C((x,w),\phi) = \{ y \in \R^3 \, :\, ||y-x||^2-w \le ||y-x'||^2-w' \quad \text{for all } (x', w') \in \phi\}, 
\end{equation} 
for $(x,w)\in\phi$.
If all weights are equal, the special case of a \emph{Voronoi  tessellation} is obtained. Laguerre tessellations are also known as \emph{power diagrams} or \emph{additively weighted Voronoi diagrams}.

The `distance' $ \pow((x,w),y) =||y-x||^2-w$ is called the {\it power} of $y$ with respect to $(x,w)$. In the case $w\geq0,$ a generator $(x,w)$ can be interpreted as a ball with center $x\in\R^3$ and radius $r=\sqrt{w}$. For points $y$ outside this ball, the power distance $\pow((x,w),y)$ measures the squared length of the tangent line from $y$ to the ball.

A further generalization can be obtained by assigning to each generator a symmetric positive definite matrix. We again retain the notation~$\phi$ for this extended set of generators, with elements denoted by $(x,M,w)$, where~$M$ is a symmetric positive definite $3\times 3$ matrix. 
Given such an extended set~$\phi$ of generators, the corresponding \textit{generalized balanced power diagram}~(GBPD)  consists of the sets 
\begin{equation}\label{eq:GPBD}
    C(\mathbf{x},\phi) = \{ y \in \R^3  :(y-x)^\top M(y-x)-w \le (y-x')^\top M'(y-x')-w' \,\text{ for all } (x',M',w') \in \phi\},
\end{equation}
for $\mathbf{x}=(x,M,w) \in \phi$. 
 Special cases of GBPDs are obtained, for example, by only considering diagonal positive definite matrices (d-GBPD) or by setting all weights $w$ to 0. In the literature, GBPDs are also referred to as {\it anisotropic power diagrams}.

\begin{figure}[h]
\centering
\includegraphics[width=0.22\textwidth]{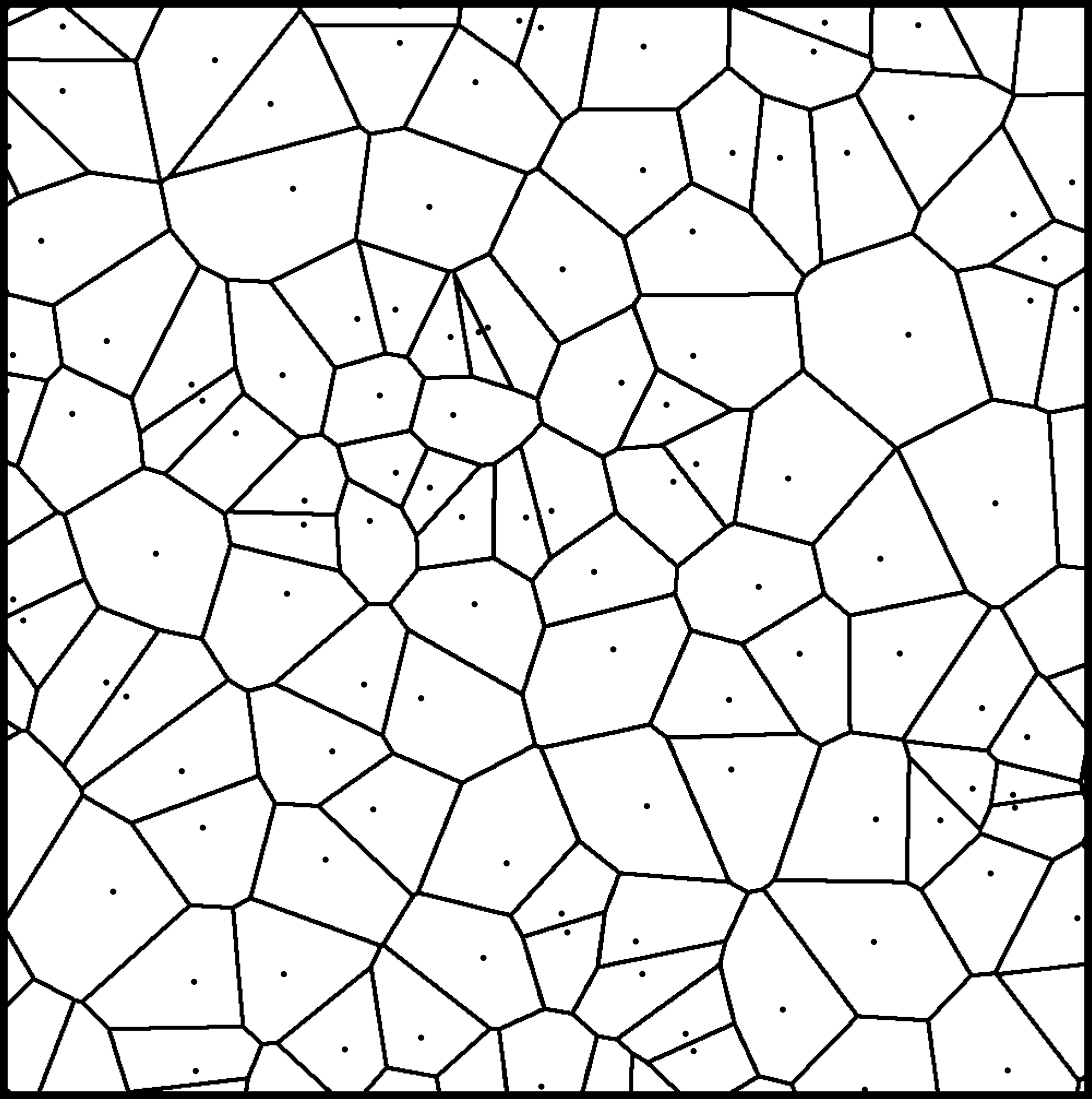}\hspace{1cm}
\includegraphics[width=0.22\textwidth]{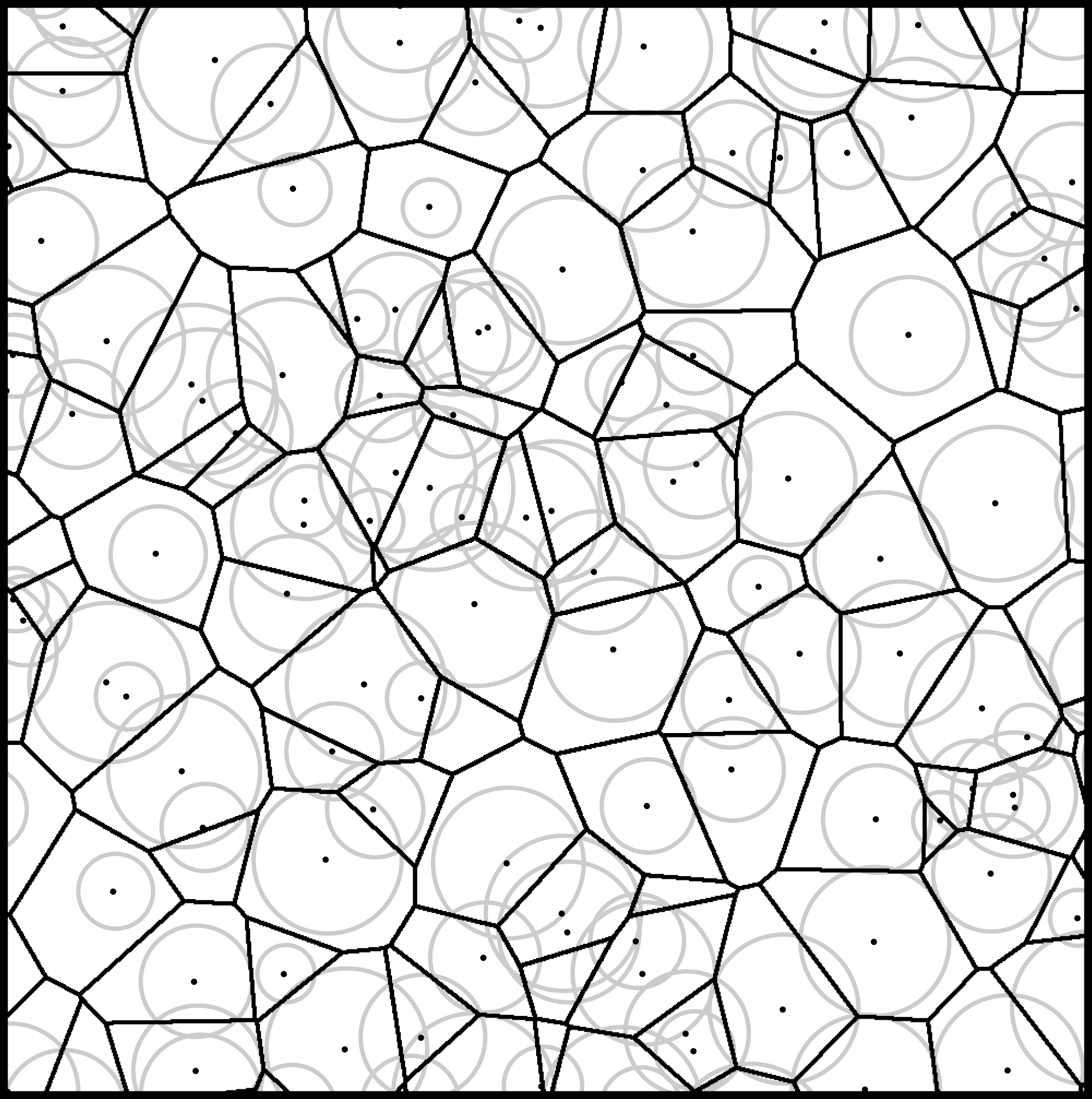}\hspace{1cm}
\includegraphics[width=0.22\textwidth]{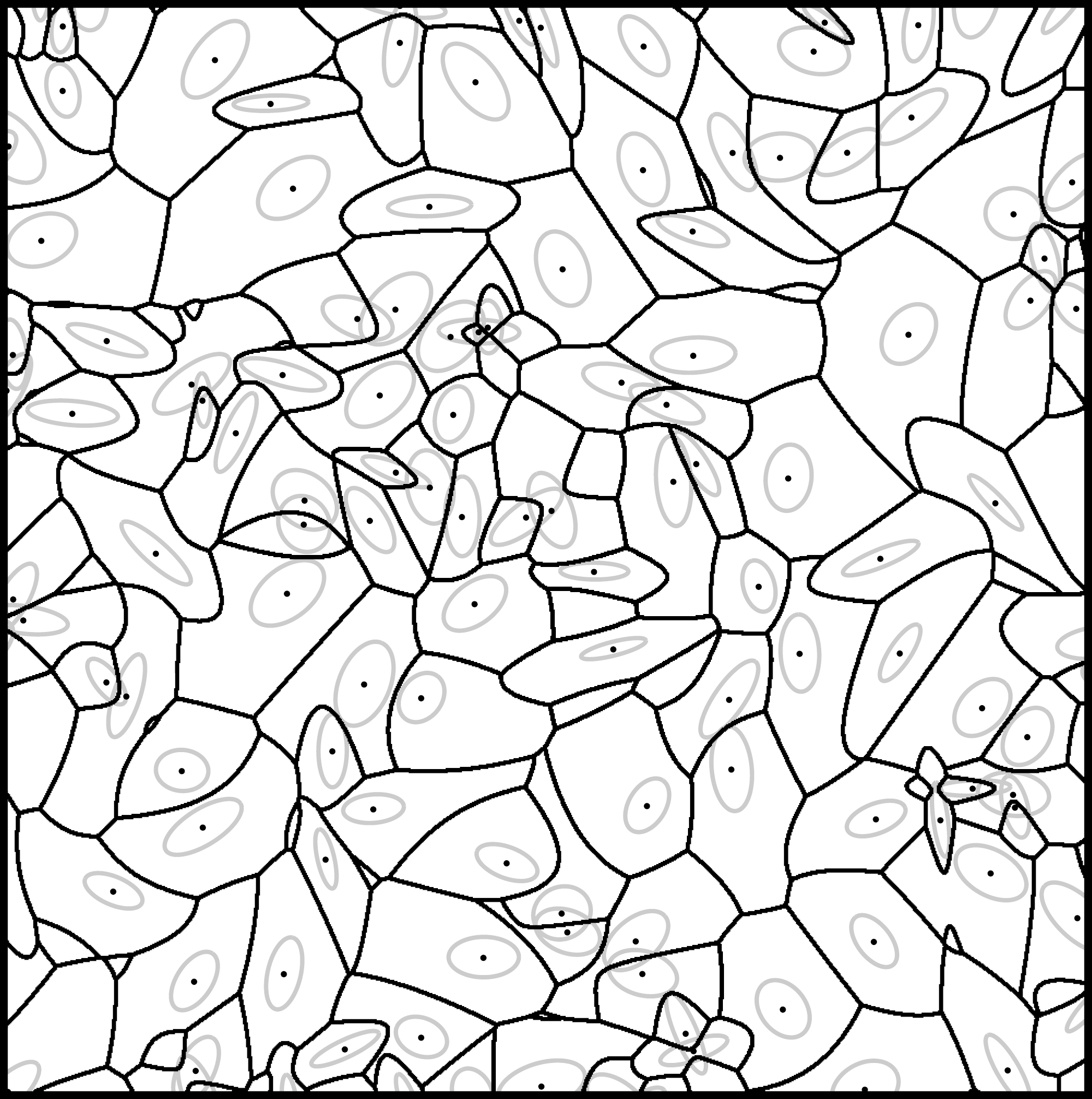}
\caption{\label{fig:Models} 
 Planar tessellations drawn from  a Poisson point process of intensity $\lambda_{\mathrm{Poi}}=100$: Voronoi tessellation (left), Laguerre tessellation (middle) with independently sampled weights $w$ (visualized by the radii of the corresponding discs) uniformly distributed on the interval [0.025,0.075], and  GBPD with independently sampled $2\times 2$ matrices $M$ representing ellipses with semi-major axes lengths uniformly distributed on [0.024,0.042], semi-minor axes lengths on [0.006,0.024] and uniform rotations on [0,$\pi$]. }
\end{figure}

Examples of planar tessellation models ($d=2$) and their generators are shown in Figure~\ref{fig:Models}, where the generator sets are drawn from a homogeneous Poisson point process in the unit square $[0,1]^2$, without marks and with appropriately chosen (independently sampled) marks. 
More information on tessellation models can be found, e.g. in~\cite{aurenhammerbook,okabebook,redenbach2025randomtessellationsoverview}.

\vspace{-0.4cm}

\section{Applications of power diagrams and their  generalizations}\label{sect:appl}

Power diagrams and their anisotropic generalizations are fundamental geometric constructs that underpin virtual materials testing and a variety of other scientific domains. The following examples illustrate their versatility and expanding range of applications, encompassing both isotropic and anisotropic cases. For comprehensive treatments of the classical (isotropic) case, see also the monographs~\cite{aurenhammerbook,okabebook,stoyanbook} and, for particular cases, the review articles~\cite{overview1,overview2,overview3}. 

\subsection{Virtual materials testing}
As outlined in Section~\ref{sec:introduction}, tessellation-based models are a useful tool to perform VMT for polycrystalline materials.
In this case, each site of the tesselation models one grain of the polycrystalline structure. For example, in~\cite{furat.2021}, tessellations were used to stochastically model the microstructure of polycrystalline particles in Li-ion battery cathodes. This approach enabled the generation of virtual polycrystalline particles based on tessellations that were subsequently used in~\cite{ALLEN2021230415} as geometric input for finite-element simulations of the particles' chemo-mechanical behavior. Such numerical simulations provided valuable insights into how grain size influences the electrochemical capacity, that is, the amount of energy that can be stored, but also into the particles' degradation behavior due to particle cracking caused by battery charge and discharge cycles. 
Beyond battery research, tessellation-based modeling has also been used to investigate the mechanical performance of further polycrystalline materials.
For example, in \cite{KOWALSKI201660} periodic Voronoi tessellations have been deployed to generate virtual three-dimensional grain architectures to represent the polycrystalline microstructure in both steels and graphite. These virtual structures served as input for  finite-element simulations of crystal plasticity, enabling the systematic study of how variations of the microstructure influence macroscopic mechanical behavior. Recently, tesselation-based models have also been used to quantify structure-property relationships of polycrystalline materials with twinning~\cite{rieder.2024, fernandes.2025}, which can be considered as a particular dependence structure between crystallographic orientations of neighboring grains.

In addition to  polycrystalline materials, tessellation models have been applied 
for the purpose of VMT of other types of materials. For example, in \cite{Jung2022} Laguerre tessellation models have been utilized to describe the cell-like pore structure within aluminum foams based on 3D image data. After fitting a model to the observed cell size distribution, virtual open foam structures were used to simulate how the elastic response changes with variations in cell shape and cross-section shape of the struts. The results quantitatively demonstrated how the geometry of the internal structure of a material influences its macroscopic mechanical properties.

\subsection{Superpixel generation}

A recent and rapidly advancing area in which (an)isotropic power diagrams play a central role is superpixel generation in imaging. Superpixels group pixels into small, perceptually homogeneous regions that respect image boundaries while reducing data complexity. Introduced by Ren and Malik in 2003~\cite{renmalik}, they have become a standard preprocessing tool in computer vision, supporting tasks such as segmentation, object detection, optical flow, and denoising \cite{app1,app2,app3,app7}.
Although early methods, such as simple linear iterative clustering \cite{slicpaper}, operate directly on the pixel grid, geometric formulations interpret superpixels as cells of a tessellation in the image plane. This shift introduces a mathematically elegant connection between image segmentation and computational geometry, allowing a more precise control over compactness, boundary adherence, and size uniformity \cite{kurlin2020,paperconvexpolygons}.

Many pixel-based methods, e.g. \cite{slicpaper,SLICmanifold,IntrinsicManifoldSLIC,vcells}, employ Voronoi diagrams as intermediate structures but depart from them to better control the sizes and shapes of superpixels. In contrast, geometric approaches such as Varane \cite{paperconvexpolygons,Duan} construct Voronoi diagrams that conform to preliminarily detected line segments, thereby producing convex-polygonal superpixels. Edge-constrained centroidal power diagrams (ECCPDs) \cite{eccpd} further generalize this framework by employing power diagrams, producing compact, resolution-independent superpixels that admit sparse representations. Although these methods yield compact, sparsely representable superpixels, they remain computationally demanding compared to pixel-based approaches. Furthermore, Power-SLIC~\cite{powerslic}, motivated by the unification and extension of these geometric frameworks through the use of anisotropic power diagrams, makes the connection between clustering formulations and representations based on tessellations explicit, thus bringing geometric methods closer to the efficiency of pixel-based techniques and suggesting that the gap between the two paradigms is beginning to close.


\subsection{Stochastic geometry}
In stochastic geometry, random elements in the space of tessellations, so-called random tessellations, are considered. 
Typically, random tessellations are generated by sets of random (marked) points, which are usually called marked point processes \cite{stoyanbook,moller2004statistical}. 
This has led to tractable stochastic 3D models for the microstructure in various materials \cite{Foehst,REDENBACH201270}.
For example, for stochastically modeling random Voronoi tessellations, it suffices to model the probability distribution of sets of unmarked points. A well-studied model is the Poisson point process, which induces the so-called Poisson--Voronoi tessellation. Various mean-value formulae exist for this and similar random tessellation models, such as explicit formulae for the mean cell volume and mean surface area~\cite{stoyanbook}. 
Besides random Voronoi tessellations, there has been a focus on random Laguerre tessellations, the generators of which are given by Poisson point processes whose points are decorated with random scalar weights---leading to Poisson--Laguerre tessellations \cite{lautensack2008random}. For this class of random tessellations, several results are available, including formulae for the volume distribution of the typical cell \cite{Gusakova2025} and estimators for the distribution of the weights, which is particularly interesting since the weights are not directly observable in Laguerre tessellations \cite{Jagt2025}. In addition to random tessellations based on Poisson point processes, in recent years so-called Gibbs--Laguerre tessellations have been explored  which introduce explicit interactions between the generating points and their weights \cite{gibbs-lag}. 
In parametric stochastic modeling, the parameters of a specific model, e.g. of a homogeneous Poisson-Voronoi tessellation, have to be determined such that the model realizations exhibit decompositions of the volume of interest that reproduce statistics of the observed grains or cells, e.g., the grain size distribution~\cite{REDENBACH20091397,VECCHIO2014171}. Furthermore, in stochastic reconstruction, a  nonparametric stochastic optimization approach is used to construct a random tessellation that fits certain predefined statistics~\cite{neper-paper,bourne.2020}. 
Recently, random tessellation models that incorporate anisotropy  have been deployed by combining anisotropic power diagrams (including GBPDs) with marked point processes. These stochastic models can capture direction-dependent grain or cell morphologies, which, in particular, can exhibit curved facets \cite{fuchs2025generating}.

\subsection{Stereology}
Stereology provides methods to infer three-dimensional geometric characteristics from lower-dimensional measurements (sections or projections) \cite{baddeley2004stereology}. 
Established results allow for the estimation of quantities such as volume fractions, surface and interface densities, and mean chord lengths for certain types of random closed sets in $\R^3$~\cite{stoyanbook}. 
Assuming that the underlying 3D geometries observed in 2D sections can be represented by power diagrams is particularly helpful in stereology, since power diagrams offer a low-parametric representation that regularizes the otherwise ill-posed inverse problem of inferring 3D geometry from 2D data.
General mean-value formulae for random 3D tessellations provide relationships that connect mean values of geometric descriptors observed in 2D sections and of the actual grains or cells of the random 3D tessellation. For example, in \cite{miles1972} explicit formulae are given that link 2D sections to mean grain/cell volumes, surface areas and face numbers.  These results demonstrate how the properties of the grains/cells in three dimensions can be  inferred from lower-dimensional observations. Recently, the distribution of 2D sections of Poisson--Voronoi tessellations have been characterized \cite{gusakova2024sectional}. These results have been extended to Poisson--Laguerre tessellations in \cite{Gusakova2025}.
 Moreover, in \cite{Jagt2025} non-parametric estimators for the weight distribution of Poisson--Laguerre tessellations have been described that can be determined from 2D sections. A stereological reconstruction of 2D cross-sections of Laguerre tessellations based on simulated annealing has been presented in~\cite{liebscher.2015}. With respect to stereological applications for anisotropic power diagrams, in \cite{fuchs2025generating} a computational approach has been presented that allows for the calibration of random GBPDs in $\R^3$ to 2D image data of planar sections by adapting methods from generative artificial intelligence. Recently, in \cite{ballani2026} it has been shown that 2D sections of GBPDs in $\R^3$ can be considered to be GBPDs in the intersecting hyperplane.


\subsection{Further examples of applications}

\emph{Data clustering and coresets.} Generalized Voronoi diagrams—including isotropic and anisotropic power diagrams—are closely related to optimal data clusterings (see, e.g.,~\cite{clusteringbriedengritzmann,clustering2,clustering3}). In particular, in~\cite{clusteringbriedengritzmann}, a one-to-one correspondence has been established between optimal solutions of certain constrained clustering problems and generalized Voronoi diagrams; see also the survey~\cite{Gr25}. Recent advances in coreset theory further enable efficient computation of such clusterings, and hence of the associated diagrams, even for very large data sets~\cite{coresets25a,coresets25b}.

\emph{Electoral district design.} The design of equitable electoral districts seeks to partition a region into compact, contiguous areas that balance prescribed population or demographic constraints. This problem can be formulated as a constrained clustering task, and geometric approaches based on generalized Voronoi constructions—particularly isotropic and anisotropic power diagrams—have been used to compute such balanced and spatially coherent partitions~\cite{electorial1,electorial2}.

\emph{Farmland consolidation.} In farmland consolidation, the goal is to reorganize fragmented agricultural plots into compact, contiguous units while preserving ownership and area constraints. This  optimization problem can be modeled as a constrained clustering task, where generalized Voronoi constructions (specifically power diagrams) yield efficient and geometrically consistent land reallocations~\cite{consolidation1,consolidation2}.

\emph{Robotic  network and service district design.}
In robotic sensing and service systems, autonomous vehicles or drones are often deployed to cover or service a geographic region while balancing spatial workloads. Such problems can be formulated as coverage optimization tasks with area or demand constraints, where the service territories correspond to generalized Voronoi tessellations. In particular, power diagrams, potentially including the anisotropic case, provide natural geometric models to determine balanced and efficient service regions~\cite{robots1,robots2}.

\emph{Mesh generation.}
In numerical simulation and scientific computing, anisotropic meshes discretize a domain by triangulations (or higher-dimensional analogs) whose elements are elongated along prescribed directions to capture directional features of the solution. Such meshes can be constructed from anisotropic Voronoi diagrams, which correspond to anisotropic power diagrams without the (or with a constant)  weight term $w$ (defined in Section~\ref{sec.tes.mod}); see~\cite{mesh2,bwy-08,canas1}.

\emph{Image stylization.} In non-photorealistic rendering, geometric constructions related to Voronoi and power diagrams have been used to emulate traditional artistic techniques. For example, centroidal Voronoi diagrams can reproduce stippling effects, where dot density conveys tone~\cite{style2}. Pebble mosaic stylization methods likewise approximate image regions by ellipsoidal shapes, conceptually related to anisotropic power diagrams~\cite{style1}.

\emph{Cell morphology modeling.}
In biological tissues, especially epithelia, the spatial organization of cells can be approximated by tessellations based on the positions of cell nuclei. The study described in~\cite{epithelium1} analyzes how well Voronoi diagrams constructed from nuclei centers reproduce actual cell shapes and neighborhood relationships observed in epithelial layers. The results provided in~\cite{epithelium1} highlight both the usefulness and limitations of such tessellations.

\emph{Deep learning and neural network architecture.}
Recent work has revealed that certain neural network architectures induce geometric structures closely related to power diagrams. In particular, networks with piecewise affine activations, such as ReLU layers, partition their input space into polyhedral regions that can be described by power diagrams. The composition of such layers yields progressive subdivisions of these diagrams, providing a geometric interpretation of how network depth increases expressive complexity~\cite{machinelearning}; see also~\cite{deepgeometry}.


\section{Optimization problems and algorithms}\label{sect:alg}
Modeling the 3D morphology of polycrystals or foams  requires finding suitable model classes that can accurately represent the grain or cell systems observed in the data. This task can be interpreted in several ways. 

\subsection{Overview of fitting approaches}\label{ove.fit.app}

{\it Fitting} or {\it approximation}, the focus of this paper, refers to the task of finding a set of generators in a given model class, e.g., a Laguerre tessellation, such that the tessellation induced by the generators gives a good fit to the
observed grain or cell system. The goodness of fit
is measured by suitable discrepancy measures. We do not make the restrictive assumption that the material must fit exactly within one of the model classes stated in Section~\ref{sec.tes.mod}, i.e., that the approximation error must be zero. When this assumption is enforced, the task is referred to as {\it inversion} in the literature~\cite{voropoly, invertingdirichlet,inverting}. 
The goal of both fitting and inversion is to generate highly accurate representations that can be directly used in downstream applications, such as automated analysis, simulation, or visualization.

\subsubsection{Types of input data}\label{typ.inp.dat}

Fitting and inversion approaches have been suggested for several types of data, i.e., for different representations of grain or cell systems. These systems/tessellations might be represented as a collection of subsets of the Euclidean space, or analytic boundary descriptions might be given. Examples of approaches that work directly with such data are proposed in~\cite{voropoly,inverting,andre}. 
The problem of approximating an arbitrary input tessellation with a Voronoi tessellation is addressed in~\cite{voropoly}, and extended in~\cite{andre} to the more general case of Laguerre tessellations. The central objective is to minimize the total mismatch volume between the grains/cells of the systems observed in the measured image data and the approximating tessellations, which is achieved via a gradient descent-based optimization method. The problem of inverting a Laguerre tessellation was solved in \cite{inverting}.

If the grain/cell system to be modeled is observed as voxel data, typically, modeling approaches voxelize the sets that make up the tessellation as well. In \cite{alpers.2015}, a fitting technique is presented by solving a linear program. It uses the relation of tessellations and optimal clusterings and formulates the problem of fitting the tessellations as a weight-balanced least-squares assignment problem. In \cite{neper-paper}, a non-linear optimization method is introduced that minimizes the distances between the observed grain/cell system and the tessellation. As an alternative, stochastic optimization methods such as simulated annealing \cite{ulm2} and a cross-entropy method \cite{spettl.2016,ulm-cross-entropy-2} were proposed. In these approaches, the generators of the tessellation are iteratively modified. Then, updated generators are accepted with a probability based on the quality of the current fit. 

 Some experimental techniques, such as X-ray diffraction microscopy~\cite{poulsenbook}, do not directly measure all grain/cell parameters;  instead, some or all of them are indirectly inferred from the measurements. For example, only volumes and centers of mass are reported in~\cite{lyckegaard.2011, alpers.2015, petrich.2021}. In this case, Lyckegard et al.~\cite{lyckegaard.2011} propose a simple heuristic to choose the generators of an approximating Laguerre tessellation. This heuristic is often chosen as initial configuration for optimization methods which then further improve the fit to real data. The approach of~\cite{alpers.2015} can also deal with such indirect data; additionally, volume limits can be incorporated into the fit. 

\subsubsection{Types of tessellations}\label{typ.tes.sel}

Approaches can also vary in terms of the tessellation types they generate. The Voronoi and Laguerre tessellations as generated in~\cite{voropoly} and \cite{lyckegaard.2011,inverting,andre}, respectively, assume that the observed structure is isotropic. However, this assumption does not always hold and individual grains can indeed exhibit strong anisotropy \cite{altendorf.2014}. In such cases, tessellations with elliptical generators such as GBPDs are superior to Laguerre tessellations \cite{SedivyModelSelection}, but require higher computational effort when fitting them to data. Some of the optimization techniques discussed above, e.g., the linear programming approaches \cite{alpers.2015,alpers.2023b}, simulated annealing \cite{ulm2} or gradient descent-based methods \cite{petrich.2021}, can also be applied to GBPDs. In the present paper, we review various algorithms for approximating the data with tessellation models of increasing complexity. Therefore, we deploy different combinations of fitting algorithms and tessellation models to several voxelized datasets, followed by an extensive quantitative comparison. The fitting algorithms were selected to cover a broad methodological spectrum, including heuristic methods, linear programming, stochastic optimization, and gradient-based methods. Within each methodological class, one or two representative methods were chosen, keeping the scope of the quantitative comparison tractable.   Additional fitting methods are discussed in Section~\ref{sect:other} to provide context and highlight further relevant developments. Table~\ref{tab:alginputouput} summarizes the  algorithms considered along with their respective inputs and outputs. 
In particular, we consider fitting algorithms based on heuristics, namely
H$_0$~\cite{altendorf.2014} and Hq~\cite{lyckegaard.2011, teferra.2018}
(Section~\ref{sec:heuristics}), and methods that rely  on linear programming (LP) as proposed
in~\cite{alpers.2015} (Section~\ref{sec:LP}) and on the cross-entropy method
of~\cite{spettl.2016} (Section~\ref{sect:CE}). Finally, we consider two methods based on gradient descent, i.e., the algorithm from \cite{petrich.2021} which we denote by GD (Section~\ref{method:gradient-descentI}) and the fitting method from the software package Neper \cite{neper-paper} (Section~\ref{sec:neper}).
The diagram types underlying these algorithms are introduced in Section~\ref{sec.tes.mod}, while the fitting algorithms are detailed in Section~\ref{sect:alg}.

\begin{table}[h!]
	\centering \small
	\begin{tabular}{@{}lcc|cccc@{}}
		\toprule
		& \multicolumn{2}{c|}{\textbf{Input Data}} & \multicolumn{4}{c}{\textbf{Output Diagram}} \\

		& \textbf{voxelized} & \textbf{indirect} & 
		\textbf{Voronoi}&\textbf{Laguerre} & \textbf{d-GBPD} & \textbf{GBPD} \\
		\midrule
		H$_0$ &\checkmark&\checkmark&\checkmark&--&\checkmark&\checkmark\\
		Hq & \checkmark  & \checkmark  & --  & \checkmark &\checkmark& \checkmark \\
		LP& \checkmark &\checkmark & --  & \checkmark &\checkmark& \checkmark\\
		CE  & \checkmark &-- & -- & \checkmark &--& -- \\
		GD & \checkmark &-- &\checkmark & \checkmark&\checkmark&\checkmark\\
		Neper &\checkmark& \checkmark &\checkmark&\checkmark&--&--\\
		\bottomrule
	\end{tabular}
	\caption{Comparison of fitting methods based on direct input data (discretized grains) and indirect input data (grain volumes and barycenters), and with respect to their 
		output data. The symbol~`\checkmark'\ in the output columns indicates that the respective diagram type can be used as output model.}\label{tab:alginputouput}
\end{table}

\subsubsection{Representation of input data as grain/cell map}\label{rep.gra.map}

 The input for all algorithms is a discretized grain/cell structure given on a grid of voxel coordinates $W = V \cap \mathbb{Z}^3$ for some set $V \subset \mathbb{R}^3$, where $\mathbb{Z}=\{\ldots, -1, 0, 1,\ldots\}$.  Throughout this paper, we will denote the number of elements in $W$ by $m\in\mathbb{N}=\{1,2,\ldots\}$. In particular, a discretized structure on $W$  
 comprising~$n\in\mathbb{N}$ grains/cells is represented by a mapping
\begin{equation}\label{rep.gra.str}
\GT : W \to \{0, \ldots, n\},
\end{equation}  
which is also referred to a \emph{grain/cell map} or  \emph{grain/cell scan}. The $i$-th grain/cell of $\GT$ is then defined by $C^{\GT}_i = \{x \in W: \GT(x) = i \}$. We assume that the cardinality $|C^{\GT}_i|$ of $C^{\GT}_i$ fulfills $|C^{\GT}_i|>1$ for each $i\in\{1,\ldots,n\}$. Typically, $\{x \in W: \GT(x) = 0 \}$ is not a grain/cell, but is either the empty set or the set of voxels that separate neighboring grains/cells.

To fit a Voronoi (or Laguerre) tessellation to a mapping $\GT$ as given in \eqref{rep.gra.str}, we consider the set of all Voronoi (or Laguerre) tessellations of $V$ that are induced by~$n$  generators. 
 In the Laguerre case, each element of this set can be represented by a (not necessarily unique) set $\phi = \{(x_1, w_1), \ldots, (x_n,w_n)\}\subseteq \R^3 \times \R$ of $n$ generators. For each $i\in\{1,\ldots,n\}$, it defines a set $C_i(\phi)= C((x_i, w_i), \phi)$ as given by Eq.~\eqref{eq:Laguerre}. The Voronoi case is then obtained by setting $w_i=0$ for each $i\in\{1,\ldots,n\}$.  For GBPDs, we use an analogous procedure where the generator sets are of the form $\phi = \{(x_1, M_1, w_1),\ldots,(x_n,M_n,w_n)\}$ and $C_i(\phi)=C((x_i,M_i, w_i), \phi)$ is given by
Eq.~\eqref{eq:GPBD}, for each $i\in\{1,\ldots,n\}$.
 For Voronoi and Laguerre tessellations as well as GBPDs, the set of all tessellations in $V$ generated by exactly $n$ generators will be denoted by~$\Phi_n$. Note that in the case of Laguerre tessellations and GBPDs, the tessellations in~$\Phi_n$ can consist of fewer than $n$ sets.

\subsection{Some heuristics [H$_0$ and Hq]}\label{sec:heuristics}
Two heuristics have been introduced in the literature that avoid optimization routines, as the generators for Voronoi and Laguerre tessellations and GBPDs are directly estimated from the data. The heuristics, henceforth referred to as H$_0$~\cite{altendorf.2014} and Hq~\cite{lyckegaard.2011, teferra.2018}, differ in their setting of the weights~$w_i,$ $1\leq i \leq n.$ In~$H_0,$ all weights $w_i$ are set to zero, which leads to a Voronoi tessellation. In contrast, under Hq, the weights are defined as
\[
w_i = \mleft( \frac{3|C_i^{GT}|}{4\pi \sqrt{\det(B_i)}} \mright)^{\!\!2/3} \!\!\!\!\!\!,  \quad  \quad i\in\{1,\ldots,n\},
\]
where the matrices $B_1,\ldots, B_n$ depend on the specific model. In the Laguerre case, the matrices $B_1,\ldots,B_n$ are taken to be the identity matrix,  as in~\cite{lyckegaard.2011}. In the GBPD case~\cite{teferra.2018}, $B_i$ is set to the sample covariance matrix $\Sigma(C_i^{GT})$ of the region $C_i^{GT},$ given by
\begin{equation}\label{cov.mat.reg}
B_i = \Sigma(C_i^{GT}) = \frac{1}{|C_i^{GT}| - 1} \sum_{y \in C_i^{GT}} (y - c(C_i^{GT})) (y - c(C_i^{GT}))^\top, \quad i\in\{1,\ldots,n\},
\end{equation}
where $c(C_i^{GT})$ denotes the barycenter of $C_i^{GT}$. 

Both heuristics, H$_0$ and Hq, have in common that the sites $x_i$ are set as $x_i=c(C_i^{GT}),$ and, in the case of GBPDs, the matrices $M_i$ are derived from a principal component analysis (PCA), where  the sample points are all  $y \in C_i^{GT}$, for each $i\in\{1,\ldots,n\}$. 
To be  precise,~$M_i$ is given as $M_i=U_i\Lambda_i^{-1}U_i^\top$ with $U_i$ denoting the $3\times 3$ matrix whose $j$-th column is the $j$-th principal component of $C_i^{GT},$ for $j\in\{1,2,3\}$. The corresponding eigenvalues $\lambda_{i1},\lambda_{i2},\lambda_{i3}$ are collected in the diagonal matrix $\Lambda_i=\textnormal{diag}(\lambda_{i1},\lambda_{i2},\lambda_{i3})$ for each $i\in\{1,\ldots,n\}$, see~\cite{teferra.2018}.

An interpretation of the heuristics within the framework of Bayesian classifiers has been given in~\cite{alpers.2023a}.

\subsection{Linear  programming based approaches [LP]}\label{sec:LP}

The linear programming approach introduced in~\cite{alpers.2015}, henceforth called LP, computes a GBPD or Laguerre diagram fitting to an observed grain structure, with the volumes of the sets $C_1(\phi),\ldots,C_n(\phi)$ lying within prescribed bounds $\kappa_1^-,\dots,\kappa_n^-$ and $\kappa_1^+,\dots,\kappa_n^+$, respectively. The bounds can either be user-defined and strictly enforced, or uniformly relaxed by assigning \(\kappa_i^- = 0\) and \(\kappa_i^+ = \infty\) for all \(i = 1, \dots, n\). GBPDs or Laguerre diagrams are generated, depending on the specification or computation of the matrices defining the ellipsoidal norms. Moreover, the approach is flexible in its input requirements, accommodating both indirect data (grain/cell volumes and barycenters), as well as direct input $C_i^{GT},$ $1\leq i \leq n.$
Here, we only consider the latter case. 

The sites $x_1,\dots,x_n\in\mathbb{R}^3$ are chosen as the barycenters of the sets $C_1(\phi),\ldots,C_n(\phi)$. The symmetric positive definite matrices $M_1,\dots,M_n\in\mathbb{R}^{3\times 3}$ are computed via a principal component analysis; more precisely,~$M_i$ is set to equal the inverse of the spatial covariance matrix of the set of voxels~$C^{\GT}_i,$ $1\leq i\leq n;$ see also Section~\ref{sec:heuristics} above. The weights $w_1,\dots, w_n\in\mathbb{R}$ defining the GBPD are finally obtained from the solution of the dual of the linear optimization problem

\begin{equation*}\label{eq:lp}
\begin{array}{lll}
\textnormal{(LP)}        &\min \sum_{i=1}^n\sum_{j=1}^m\gamma_{i,j}\xi_{i,j}     &\\
\textnormal{subject to}  &\sum_{i=1}^n\xi_{i,j}=1                                 &(1\leq j \leq m),\\
                         &\kappa_i^-\leq\sum_{j=1}^m \xi_{i,j} \leq \kappa_i^+               &(1\leq i\leq n),\\
                         &\xi_{i,j}\geq 0                                         &(1\leq i\leq n;\: 1\leq j \leq m),
\end{array}
\end{equation*}
where $\gamma_{i,j}= (y_j-x_i)^\top M_i (y_j-x_i)$ and $y_j$ denotes the position of voxel~$j$  for all $i\in\{1,\ldots,n\},j\in\{1,\ldots,m\}$. The $\xi_{i,j}$ are the variables; they specify the fraction of voxels $y_j$ that are assigned to site~$x_i$. In the optimum, it can be ensured that these fractions are, in fact, binary due to the special structure of the linear program. 

For the results presented in this paper we used the common choice of setting the volume bounds to $\kappa_i^-=\kappa_i-\varepsilon$ and $\kappa_i^+=\kappa_i+\varepsilon$ with $\varepsilon=2$  and $\kappa_i$ denoting the volume of $C^{\GT}_i.$ No initial solution is required.

The above LP approach, through its dual formulation,  optimizes only the weights $w_1, \dots, w_n$. In contrast, the following method from~\cite{alpers.2023b}, framed within a support vector machine context, optimizes all parameters—that is, \( x_1, \dots, x_n \), \( M_1, \dots, M_n \), and \( w_1, \dots, w_n \)—by solving a linear program. (As before, the weights can be obtained via the dual formulation.) However, this approach does not incorporate volume bounds.  
 Using the notation
\begin{equation*}
    y_j= \left(
    \begin{array}{c}
         (y_j)_1  \\
         (y_j)_2 \\
         (y_j)_3 \\
    \end{array}
    \right), \qquad
    M_i= \left(
    \begin{array}{ccc}
         (M_i)_{1,1} & (M_i)_{1,2} & (M_i)_{1,3}  \\(M_i)_{1,2} & (M_i)_{2,2} & (M_i)_{2,3}  \\ (M_i)_{1,3} & (M_i)_{2,3} & (M_i)_{3,3}
    \end{array}
    \right),
\end{equation*}
and setting
$
    a_{i} = -2\apdmatrix_{i}x_{i}$ and $\alpha_{i} = x^{\top}_{i}\apdmatrix_{i}x_{i}+\apdweight_{i}$,
all parameters are encoded in the extended parameter vector
$
	\fa_{i} = \bigl(\alpha_{i},a_{i}^{\top},(\apdmatrix_{i})_{1,1},2(\apdmatrix_{i})_{1,2}, 2(\apdmatrix_{i})_{1,3}, (\apdmatrix_{i})_{2,2},2(\apdmatrix_{i})_{2,3},(\apdmatrix_{i})_{3,3}\bigr)^{\top}\in \R^{10}$ which yields the variables of the linear program. Setting \begin{equation*}
	\fx_j = \bigl(1,(y_j)_{1},(y_j)_{2},(y_j)_{3},(y_j)_{1}^{2},(y_j)_{1}(y_j)_{2},(y_j)_{1}(y_j)_{3},(y_j)_{2}^{2},(y_j)_{2}(y_j)_{3},(y_j)_{3}^{2}\bigr)^{\top},
\end{equation*} the linear program is then given by

\begin{equation*}
\begin{array}{ccrclcl}
	&\multicolumn{4}{c}{\displaystyle \min_{(\zeta_j)} \quad \sum_{j=1}^{n}\zeta_j}&& \\[.6cm]
	&&	\displaystyle \fa_{i}^{\top}\fx_j - \fa_{\ell}^{\top}\fx_j +1  & \le  & 0 &\quad & \bigl(j\in \{1,\dots,m\}, \, i,\ell \in \{1,\dots,n\}, \, \ell\ne i,\, y_j\in \operatorname{int}_{\delta_i}(C_i) \bigr), \\[.2cm]
	&&	\displaystyle \displaystyle \fa_{i}^{\top}\fx_j - \fa_{\ell}^{\top}\fx_j - \zeta_j & \le  & 0 &\quad & \bigl(j\in \{1,\dots,m\}, \, i,\ell \in \{1,\dots,n\}, \, \ell\ne i, \, y_j\in C_i\setminus \operatorname{int}_{\delta_i}(C_i)\bigr), \\[.2cm]
	&&	              \zeta_{j} & \ge  & 0          &\quad & \bigl(j \in \{1,\dots,m\}, \, y_j\in C_i\setminus \operatorname{int}_{\delta_i}(C_i)\bigr),\\
\end{array}
\label{lp:polynomial_boundaries_II}
\end{equation*} where $\operatorname{int}_{\delta_i}(C_i)$ denotes the $\delta_i$-interior of $C_i,$ which is defined by $\operatorname{int}_{\delta_i}(C_i)=\{x\in C_i: ||x-x'||\geq \delta_i\: \textnormal{for all } x'\not\in C_i\},$ for user-specified  parameters $\delta_i,$ $1\leq i \leq n.$

To accelerate computations and reduce memory usage, the calculations for both approaches may be restricted to voxel subsets forming a so-called \emph{coreset}, as described in~\cite{alpers.2023b} (see~\cite{fiedler2024resolution} for the mathematical foundations and rigorous proofs). Following the procedure in~\cite{alpers.2023b}, we sample every 10\textsuperscript{th} voxel along the $x$-, $y$-, and $z$-directions within each input set, excluding their $\delta_i$-interior regions, with $\delta_1 = \cdots = \delta_n = 20$.

It should be noted that, in contrast to the other fitting methods discussed in this paper, LP generates a GBPD that adheres to user-specified volume constraints. Remarkably, however, it does not explicitly minimize a discrepancy with respect to the ground truth structure GT within its objective function. The fit appears to reflect the model’s capacity to approximate physical processes, for example, those that govern the formation of polycrystalline structures.

\subsection{Fitting with the
 cross-entropy method [CE]}\label{sect:CE}

In this section, we summarize the method introduced in~\cite{spettl.2016} for fitting Laguerre tessellations to grain structures $\{C_i^{GT}: 1\leq i \leq n \}$ in polycrystalline materials by minimizing an interface-based discrepancy measure, using the cross-entropy method~\cite{deboer.2005},  henceforth called CE. 
For this purpose, we assume that $V \subset \R^3$ is connected. For any $i,j\in\{1,\ldots, n\}$ with $i\not= j$, the discrete interface between two grains $C^{\GT}_i$ and $C^{\GT}_j$ is defined by
\begin{equation*}
    N^{\GT}_{i,j} = \{ x \in W: \mathcal{N}_{26}(x) \cap C^{\GT}_i \neq \emptyset \textnormal{ and }\mathcal{N}_{26}(x) \cap C^{\GT}_j \neq \emptyset \},
\end{equation*}
where $\mathcal{N}_{26}(x)$ denotes the $26$-neighborhood of $x$ in $W,$  consisting of all voxels that share a face, edge, or vertex with $x$; see Section~3.3 in~\cite{ohser.2009}. Note that $N^{\GT}_{i,j} = \emptyset$ if and only if $C^{\GT}_i$ and $C^{\GT}_j$ are not adjacent. This definition of adjacency means that two grains are also adjacent if they are separated by a line that is one voxel thick. This is a reasonable definition for the original data in \cite{spettl.2016}, where all grains are separated by boundaries that have a thickness of one voxel. Even for data without such separating voxels, the definition of $N^{\GT}_{i,j}$ is unproblematic since grains that are only one voxel thick do usually not appear in applications. 

Based on this notion of neighborhood voxels, we introduce the interface-based discrepancy measure $E: \Phi_n \rightarrow [0,\infty)$ by 
\begin{equation}\label{eq:costfunction_CEmethod}
    E(\phi) = \sum_{i=1}^{n-1} \sum_{j = i+1}^n \sum_{x \in N^{\GT}_{i,j}} \min \{||x-y||^2: y \in C_i(\phi) \cap C_j(\phi)\}.
\end{equation}
Note that the cells \(C_i(\phi)\) are closed sets, i.e., for neighboring cells
\(C_i(\phi)\) and \(C_j(\phi)\), the intersection in Eq.~\eqref{eq:costfunction_CEmethod}
is non-empty and consists of their shared boundary (interface) points. 
 The discrepancy measure~$E$ given in Eq.~\eqref{eq:costfunction_CEmethod} quantifies the distance between the discretized grain boundaries of the input data and the cell boundaries of the tessellation of~$\phi$. Minimizing the value of~$E(\phi)$ constitutes a high-dimensional optimization problem, typically characterized by numerous local minima, and the evaluation of the discrepancy measure~$E$ is computationally expensive. Thus, it is proposed in ~\cite{spettl.2016} to use the cross-entropy method to minimize an approximation of $E$, where---instead of considering all points $x \in N^{\GT}_{i,j}$---only a smaller set of test points $x \in T_{i,j} \subset N^{\GT}_{i,j}$ is considered for any $i,j\in\{1,\ldots,n\}$ with $i< j$, see Section~3.4 of~\cite{spettl.2016}. 

Note that the cross-entropy method is a stochastic optimization method that can be used for parameter fitting~\cite{rubinstein.2004}. In our case, the parameter vector consists of all generators of the Laguerre tessellation. For a predefined initial parameter vector, the discrepancy measure~$E$ is evaluated for a random sample of 4,000 parameter vectors. Those are assumed to be normally distributed around the initial parameter vector with some variance. Then, the elite set, that is, those 200 parameter vectors for which~$E$ takes the smallest values, is selected. The parameter vector and variance are then updated to equal the mean and variance, respectively, of the parameter vectors in the elite set. This procedure is iterated until the cost function does not decrease significantly for a given number of steps. As an initial parameter vector for the cross-entropy method, the generators obtained from the heuristic Hq are used; see Section~\ref{sec:heuristics}. The authors of~\cite{spettl.2016} made the code for this fitting approach publicly available~\cite{laguerreUlmcode}. 

In principle, the interface-based discrepancy measure considered in~\cite{spettl.2016} can also be used to fit GBPDs. However, in this scenario, the computation of the minimum distance to the boundary $C_i(\phi) \cap C_j(\phi)$ is significantly more challenging, rendering the method computationally demanding. Note that in the case of Laguerre tessellations, $C_i(\phi) \cap C_j(\phi)$ is contained in an affine subspace, such that the analytical determination of the minimum distance given in Eq.~\eqref{eq:costfunction_CEmethod} can be carried out efficiently.

\subsection{Gradient descent-based fitting I [GD]}\label{method:gradient-descentI}

We now consider an algorithm which is based on gradient descent methods, henceforth called GD.
For this, in \cite{petrich.2021}, the goodness of fit between the discretized grain/cell structure $\GT$ and a GPBD with generator $\phi$ is measured using the volume-based measure $E\colon \Phi_n \to \R$ given by
\begin{equation}\label{UUGD:objective}
E(\phi) = \frac{1}{|W|} \sum_{x \in W} \sum_{i=1}^n \lambda\!\left( \mathbbm{1}_{C_i^\GT}(x), \mathbbm{1}_{C_i(\phi)}(x)  \right),
\end{equation}
where $\lambda(y,y^\prime) \in \R$ measures the similarity between pairs of values $y,y^\prime \in [0,1]$,  and $\mathbbm{1}_A$ denotes the indicator of the set $A \subseteq W$, defined by $\mathbbm{1}_A(x) = 1$ if $x \in A$, and $\mathbbm{1}_A(x) = 0$ if $x \in W\setminus A$.

Note that even if $\lambda$ is differentiable in the second argument, the function $E$ given in~\eqref{UUGD:objective} is not differentiable with respect to the GBPD generators $\phi$. Hence, maximization of $E$ based on gradient descent methods is not viable. This problem can be solved by approximating $E$ with a differentiable objective function. For this purpose, the so-called one-hot encoding was employed in~\cite{petrich.2021}.

First, let $D \colon W \times  \Phi_n  \to \R^n$ be given by
$D(x,\phi) = \left( (x-x_i)^\top M_i (x-x_i) -w_i \right)_{i=1}^n,$
i.e., the $i$-th component of the vector $D(x,\phi)$ is the `distance' of $x$ to the generator point $\mathbf{x}_i=(x_i,M_i,w_i)$, for each $i\in\{1,\ldots,n\}$. Since $x \in C_i(\phi)$ holds whenever the ‘distance’ from $x$ to $\mathbf{x}_i$ is less than or equal to the distance to any other generator point, the second argument of $\lambda$ in~(\ref{UUGD:objective}) can alternatively be written as
\begin{equation}\label{UUGD:onehot}
\mathbbm{1}_{ C_i(\phi)}(x) = \argmin_i^*(D(x, \phi)),
\end{equation}
where \( \argmin_i^* \) denotes the \( i \)-th component of the function
$
\argmin^* \colon \mathbb{R}^n \to \{0,1\}^n,$
which maps a vector $z = (z_1, \dots, z_n) \in \mathbb{R}^n$ to a binary vector indicating the locations of its minimal components. Specifically,
\[
\argmin_i^*(z) =
\begin{cases}
1, & \text{if } z_i = \min\{z_1, \dots, z_n\}, \\
0, & \text{otherwise},
\end{cases}  
\]
for each $ i\in\{1,\ldots,n\}.$ 
That is, $\argmin^*_i(z)$ is the one-hot (or multi-hot, in case of ties) encoding of the set of indices at which the minimum of \( z \) is attained. Note that $\argmin_i^\ast$ is not differentiable everywhere, which transfers also to Eq.~(\ref{UUGD:onehot}) and consequently to Eq.~(\ref{UUGD:objective}). To address this issue, in \cite{petrich.2021}, a procedure is proposed to approximate the vector-valued argmin function by the softmax function with negated inputs, $\softmax^* \colon \mathbb{R}^n \to [0,1]^n,$
whose $i$-th component is defined as  
\[
\softmax_i^*(z) = \frac{\exp(z_i)}{\sum_{j=1}^n \exp(z_j)},
\]
for any $z = (z_1, \dots, z_n) \in \mathbb{R}^n$ and $i\in\{1,\ldots,n\}$.  
In other words,  
$
\argmin^*(z) \approx \softmax^*(-z)$  for each $z = (z_1, \dots, z_n) \in \mathbb{R}^n$. Thus, a differentiable approximation $\widetilde{E} \colon \Phi_n \to \R$ of the objective function $E$  to be maximized is given by 
\begin{equation}\label{UUGD:objectiveApproximative}
\widetilde{E}(	\phi) = 
\frac{1}{|W|} \sum_{x \in W} \sum_{i=1}^n 
\lambda\!\left(  \mathbbm{1}_{C_i^\GT}(x) ,
	\softmax_i^\ast\left( -D(x,\phi) \right)
\right).
\end{equation}
In particular, in~\cite{petrich.2021}, the similarity measure 
$\lambda \colon \{0,1\}\times [0,1] \to (-\infty,0]$ given by the negative binary cross-entropy loss  
$
\lambda(y,y^\prime) = y \log y^\prime + (1 - y) \log (1 - y^\prime)$ is employed, 
where $y\in \{0,1\}$ and $y^\prime \in [0,1]$. Finally, to obtain the fitted GBPD model, the negative objective function $-\widetilde{E}$ is minimized using a (GPU-accelerated) stochastic gradient descent algorithm and running it for up to 25 iterations.

\subsection{Gradient descent-based fitting II [Neper]}\label{sec:neper}
In \cite{neper-paper}, fitting Laguerre tessellations to data is formulated as a non-linear optimization problem, where the optimization variables are the coordinates and weights of the generators. For $n$ generators, this leads to a total of $4n$ variables.
	The approach, henceforth called Neper with reference to its implementation in the Neper package \cite{neper-software}, was developed to account for statistical reconstruction of grain/cell sizes and shapes. However, it can also be used for a grain/cell-wise reconstruction based on microscopy measurements or 3D image data. Here, we use the latter.

	The aim is to minimize the discrepancy between the measured grains/cells and the sets $C_i(\phi)$ of a Laguerre tessellation.   To this end, for each $i\in\{1,\ldots,n\}$, let
\[
\partial C_i^{\mathrm{GT}} = \{ x \in C_i^{\mathrm{GT}} :
\text{there is } y \in \mathcal{N}_6(x) \text{ with } y \notin C_i^{\mathrm{GT}} \}
\]
denote the boundary voxels of the cell $C_i^{\mathrm{GT}}$ with $\mathcal{N}_6(x)$ denoting the set of voxels that share a face with~$x$. Again, it is useful to assume  that $V\subset \mathbb{R}^3$ is connected. The discrepancy of a grain/cell $C^{\GT}_i$ and its corresponding Laguerre set $C_i(\phi)$ is measured by the sum of squared Euclidean distances from the points in $\partial C_i^{\GT}$ to the set $C_i(\phi).$ This yields  the objective function 
	\[\mathcal{O}(\phi)=\frac{2}{\bar{d}\sum_{i=1}^n|\partial C_i^{\GT}|} \sum_{i=1}^n \sum_{x \in \partial C_i^{\GT}} \inf \{||x-y||^2: y \in C_i(\phi)\},\]
 where $\bar{d}$ denotes the mean grain/cell diameter in the grain/cell map GT.

Minimizing \(\mathcal{O}(\phi)\) with respect to the variables \(x_1, \dots, x_n, w_1, \dots, w_n\) defines an unconstrained non-linear optimization problem. The implementation in Neper employs the C++ NLopt library \cite{NLopt} to solve the problem using derivative-free methods: Subplex \cite{neper-subplex}, a variant inspired by the Nelder--Mead simplex algorithm, and Praxis \cite{neper-praxis}, which performs line minimization along principal axes. The initial parameter values are selected according to the heuristic Hq (see Section~\ref{sec:heuristics}).

The optimization can terminate based on several criteria; following \cite{neper-paper}, we adopt an absolute error criterion, stopping when the improvement in the objective function over \(40n\) iterations falls below \(10^{-3}\).

\subsection{Other algorithms from the literature}\label{sect:other}

\subsubsection{Gibbs-Laguerre tessellations}\label{gibbs-laguerre}In~\cite{gibbs-lag}, the reconstruction problem is addressed by generating a realization of an appropriate Gibbs point process. The point process distribution is governed by an energy function designed to minimize discrepancies between the sets of a Laguerre tessellation and the target distribution of grain/cell characteristics. Various methods to incorporate correlations between these characteristics have also been developed \cite{schladitz-particles}. Although this approach has mainly been applied to stochastic reconstruction, it can be adapted to minimize discrepancies in  grain/cell terms. However, since the method is designed to match the geometric features of polyhedra, it does not readily apply to the reconstruction of voxel-based data.

\subsubsection{Gradient descent-based polyhedra matching}

A further gradient descent-based method is described in \cite{andre}. Similarly to the method mentioned in Section~\ref{gibbs-laguerre}, it relies on the polyhedral representation of the sets $C_i(\phi)$ of a Laguerre tessellation. The objective is to minimize the total mismatch volume between the sets $C_i(\phi)$ and the observed grains/cells. The total mismatch volume can be expressed in terms of a continuously differentiable function which needs to be minimized. The variables correspond to the set of generators.

\subsubsection{Optimal transport-based approaches}

An optimal transport--based approach for generating Laguerre diagrams with prescribed  volumes of the sets $C_i(\phi)$ has been introduced in~\cite{bourne.2020}. This approach is based on the result that for any given set of volumes, a corresponding Laguerre tessellation exists, and finding such a tessellation reduces to solving a convex optimization problem. Although the primary focus is not on fitting, the authors of~\cite{bourne.2020} demonstrate that the method can be adapted for this purpose by initializing the seed points at the centroids of the target sets $C_i(\phi)$. The subsequent work~\cite{bourne2025inverting} proposes a method for generating a Laguerre diagram or GBPD with prescribed volumes whose centroids are close---in the least-squares sense---to a given set of target centroids. Furthermore,~\cite{buze-pyAPD} presents a highly efficient, GPU-accelerated implementation of a semidiscrete optimal transport method, accompanied by a Python library, for generating GBPDs with prescribed statistical properties.

\subsubsection{Simulated annealing approaches}

The algorithm described in \cite{ulm2} and \cite{SedivyModelSelection} is based on the simulated annealing optimization method, which draws inspiration from the physical annealing process.  Starting from an initial parameter configuration, it iteratively perturbs a single parameter associated with a randomly selected generator, thereby producing an updated discrepancy value. The lower the new discrepancy, the higher the probability of the new parameter configuration being accepted. The speed of convergence can be controlled via a cooling schedule that is included in the acceptance probabilities.

\subsection{Some general considerations} All methods discussed above fit a given grain/cell structure using a tessellation from a specified model class introduced in Section~\ref{sec.tes.mod}. The quality of the fit is assessed---either directly or indirectly, as in the LP approach---using an appropriate discrepancy metric, which is subsequently minimized with respect to the set of generators. For solving the optimization problem, iterative approaches such as the cross entropy method or gradient descent as well as a direct linear program can be used.  
 In the literature, both volume-based and interface-based measures are considered, see the discussion in \cite{spettl.2016}. Volume-based measures aim at maximizing the volume overlap between the observed grains/cells and their counterparts in the tessellation approximation. In contrast, interface-based measures only consider fitting the grain/cell boundaries. Interface-based measures are more efficient from a computational point of view since only boundary voxels need to be checked. For Laguerre tessellations, the measures can additionally be evaluated analytically without discretizing the tessellation. For GBPDs this is currently not possible in 3D as no explicit representation of the boundaries of the sets $C_i(\phi)$ is known, see \cite{JUNG2024102101} for the 2D case. To reduce the runtime further, the computation of the discrepancy measure can be reduced to a subset of voxels. This approach is applied by the CE method for the interface-based method and by LP when using the so-called \emph{coresets}~\cite{alpers.2023b} in a volume-based approach.

In general, designing a fitting method can be viewed as a modular task that involves selecting (a) a class of tessellation models, (b) a data representation format, (c) a discrepancy measure, and (d) an optimization scheme—typically with associated hyperparameters. The performance of the optimization step can be further enhanced by specifying an appropriate initial configuration and, in many cases, a stopping criterion.

\section{Image data}\label{sect:data}

We evaluate several 3D polycrystal and foam image data sets, which represent realistic examples from materials science, to assess the performance of the algorithms stated in Section~\ref{sect:alg}.

\subsection{Polycrystals}
\subsubsection{Time-resolved microstructure of AlCu polycrystals}

\begin{figure}[h]
    \centering
    \begin{tabular}{cccc}
         \includegraphics[width = 0.2 \textwidth]{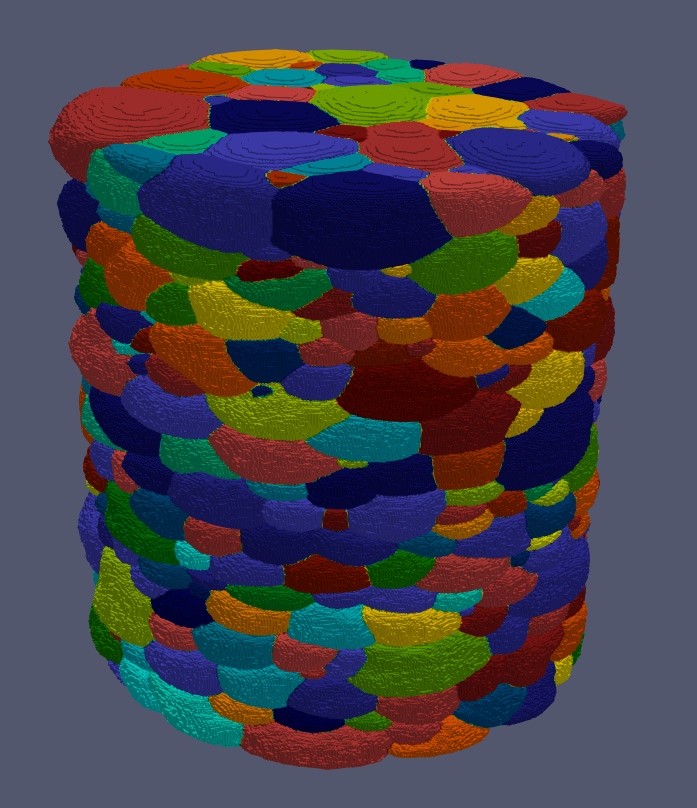} &
\includegraphics[width = 0.24 \textwidth]{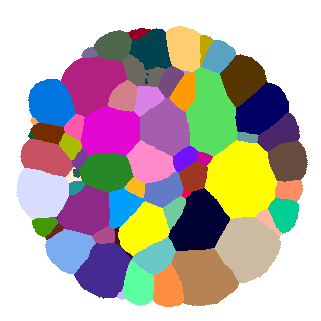} &
\includegraphics[width = 0.24 \textwidth]{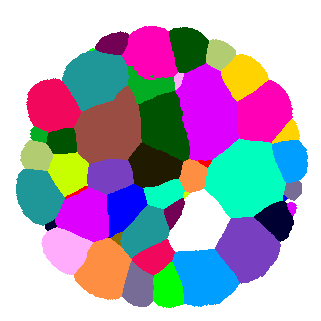} & \includegraphics[width = 0.24 \textwidth] {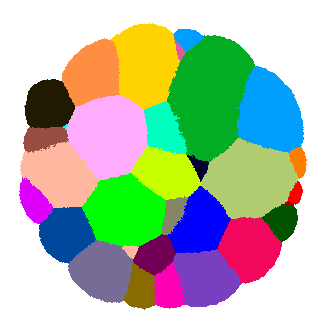} 
    \end{tabular}
    
    \caption{Image data representing grain in AlCu polycrystals, provided by C. E. Krill III. A~3D rendering at the initial time step (left), as well as 2D cross-sections after time steps $0,\,$ $20,\,$ and $60\,$ ~(center left, center right, right) are shown. The image size is $531 \times 321 \times 321$  with a voxel edge length of \SI{5}{\micro\meter}. Note that due to the non-convex shape of grains, they can exhibit multiple connected components within 2D cross-sections. }
        \label{fig:DataAlCu}
\end{figure}

We consider time-resolved 3D image data representing the polycrystalline microstructure of an aluminum (Al) copper (Cu) alloy. 3D imaging of a  $1.4~\mathrm{mm}$-diameter cylinder of
Al-5~wt$\%$Cu has been performed by means of 3DXRD measurements as described in Section~3.2 of~\cite{furat.2019}. Image data has been acquired for seven different time steps during grain coarsening by Ostwald ripening. Each 3D image has a size of $531 \times 321 \times 321$ voxels, where the cubic voxels have an edge length of \SI{5}{\micro\meter}. In the present paper, we focus on the data corresponding to the initial time step $t=0,\,$ as well as on data for  $t=20,\,$ and $t=60.\,$ At the initial time, the image comprises~938 grains. Off-grid representations of the grain system by parametric tessellations such as Laguerre tessellations and GBPDs are discussed in~\cite{petrich.2021}, where the method described in Section~\ref{method:gradient-descentI} has been used to fit tessellation models to image data. Slices and a 3D rendering of the AlCu polycrystal data are shown in Figure \ref{fig:DataAlCu}. We will refer to these data sets as \textbf{AlCuStep0}, \textbf{AlCuStep20}, and \textbf{AlCuStep60}.

\vspace{-0.4cm}

\subsubsection{Time-resolved microstructure of 99.9\% pure iron polycrystals}\label{sec:DataAA}
\begin{figure}[h]
    \centering
    \includegraphics[width=0.3\textwidth]{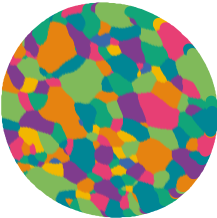}
    \caption{2D  slice through the 3D dataset of 99.9\% pure iron~\cite{alpers.2023a}, provided by H. F. Poulsen. The image size is  \(280 \times 320 \times 256\) with a voxel edge length of~\SI{1.54}{\micro\meter}.}
    \label{fig:DataPureIron}
\end{figure}

We also consider time-resolved 3D image data representing the polycrystalline microstructure of a 99.9\% pure iron sample. The data set is discussed and analyzed in detail in~\cite{Zhang.2020, Zhang.2018}, where the complete data processing workflow—from the initial experiment to the acquisition of grain scans—is thoroughly described, particularly within the supplementary materials. The data set, obtained by a diffraction contrast tomography (DCT) \cite{dct1,dct2} experiment at beamline ID11 at the European Synchrotron Radiation Facility~(ESRF), comprises 15 3D images, which were taken at different time steps of an annealing process. The material was first cold-rolled and annealed for~30 minutes at 700\textdegree C to fully recrystallize before it was first scanned. In the present paper, we focus on the data for the initial time step $t=0.\,$ At this time step, the \(280 \times 320 \times 256\) voxel image, composed of cubic voxels each with an edge length of~\SI{1.54}{\micro\meter}, contains a total of 1,327 grains. A 2D slice through this data is shown in Figure~\ref{fig:DataPureIron}. We will refer to the data set as \textbf{PureFeStep0}.

\vspace{-0.4cm}

\subsection{Foams}
\subsubsection{Ceramic foam}
The first foam sample is a ceramic foam with a size of $\SI{2.1}{\centi\meter} \times \SI{5}{\centi\meter} \times \SI{5}{\centi\meter}.$ The pore size is 20 ppi (pores per inch). The sample was scanned by $\mu$CT at the Fraunhofer ITWM in Kaiserslautern, Germany, with a voxel edge length of~$\SI{33.91}{\micro\meter}$ resulting in an image size of approximately $700 \times 1550 \times 1500$ voxels.

\begin{figure}[h]
    \centering
    \includegraphics[width=0.52\textwidth]{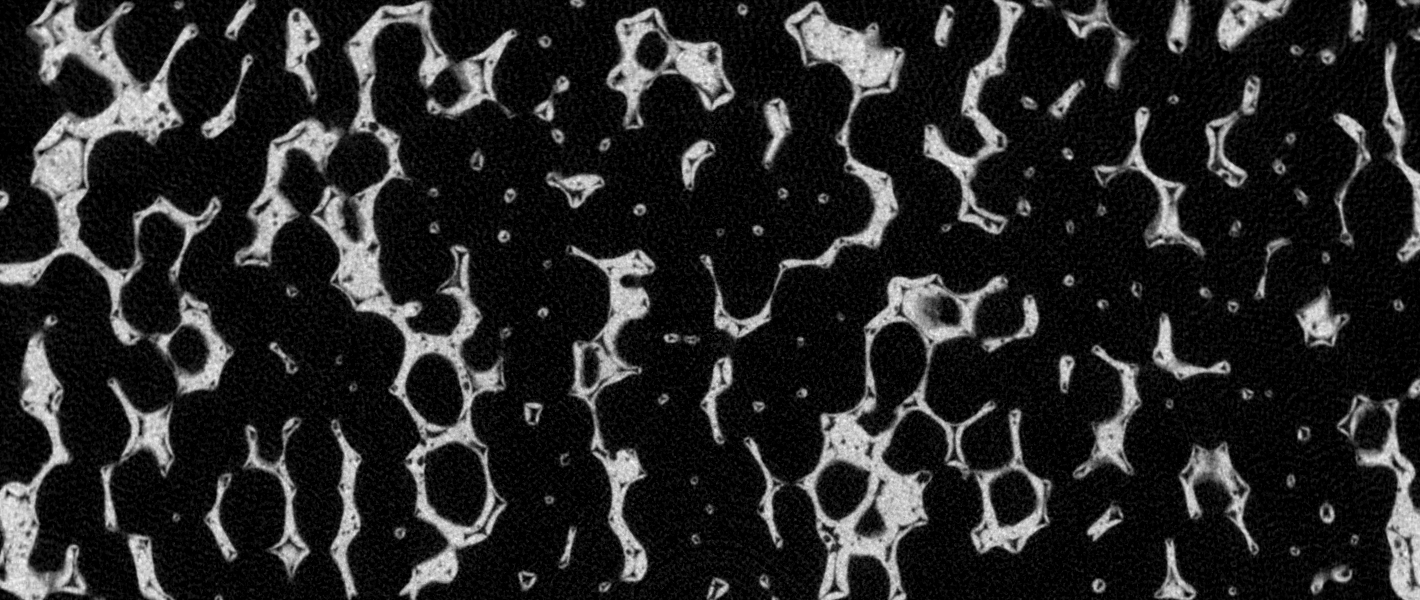}\\
    \vspace{3px}
    \includegraphics[width=0.52\textwidth]{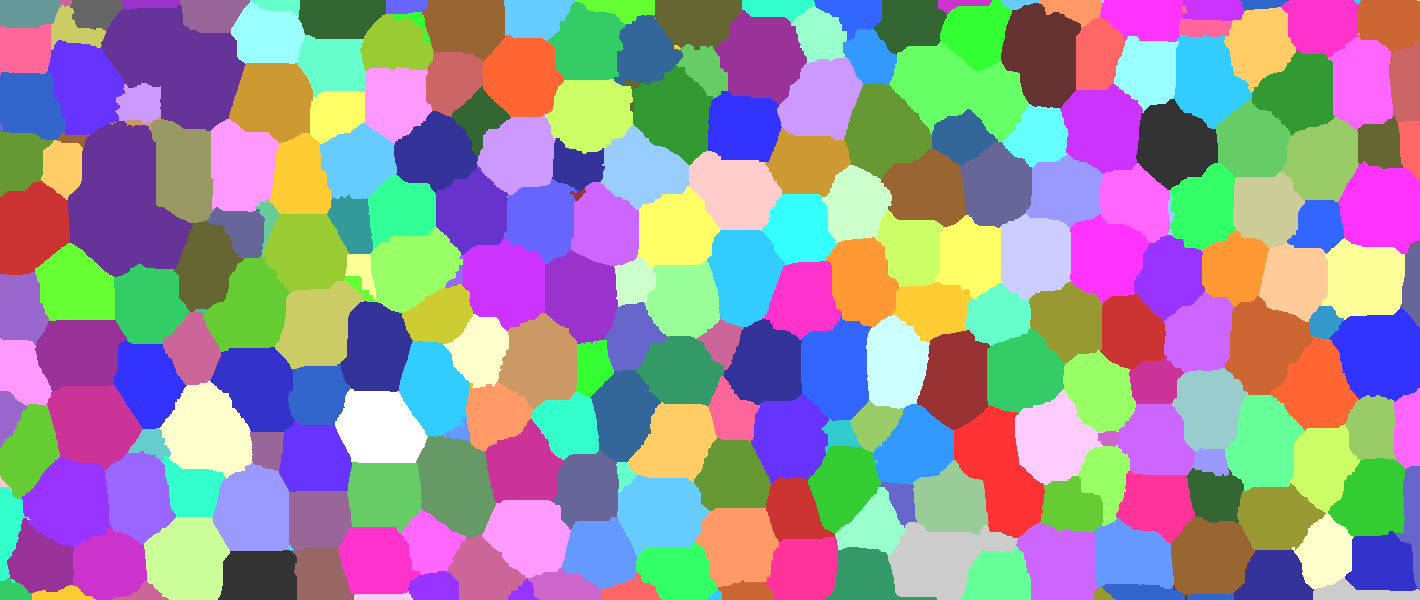}\\
    \includegraphics[width=0.52\textwidth]{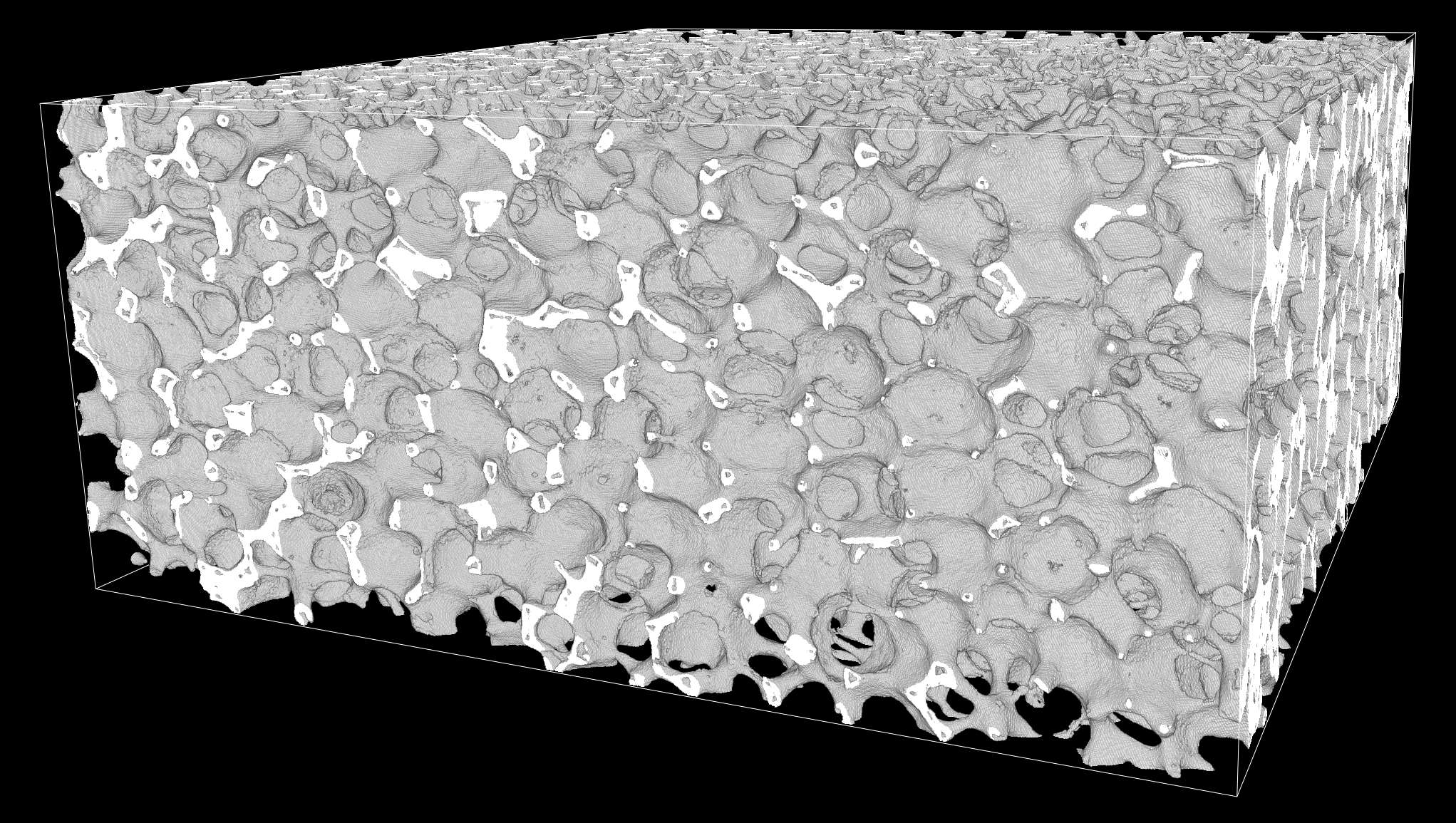}
    \caption{2D slice of the reconstructed CT image of a partially closed ceramic foam (top), cell reconstruction by watershed (middle), and a volume rendering (bottom). The image size is $700\times 1550\times 1500$ voxels with a voxel edge length of~\SI{33.91}{\micro\metre}\label{fig:CeramicFoam}.}
\end{figure}

The wall system of this partially closed foam was segmented and analyzed in \cite{kampf2015}. Stochastic models for the microstructure were presented in \cite{lautensack07:_model,RedWirRie09}. A 2D slice of the ceramic foam image along with its segmentation is shown in Figure \ref{fig:CeramicFoam}. Here, we consider a cropped section of the image of size $ 400\times 800\times 800$ voxels that contains 1,003 pores. We will refer to the data set as \textbf{Ceramic}.

\subsubsection{Closed polymer foam}

\begin{figure}[h]
     \centering
   \includegraphics[height=0.25\textwidth]{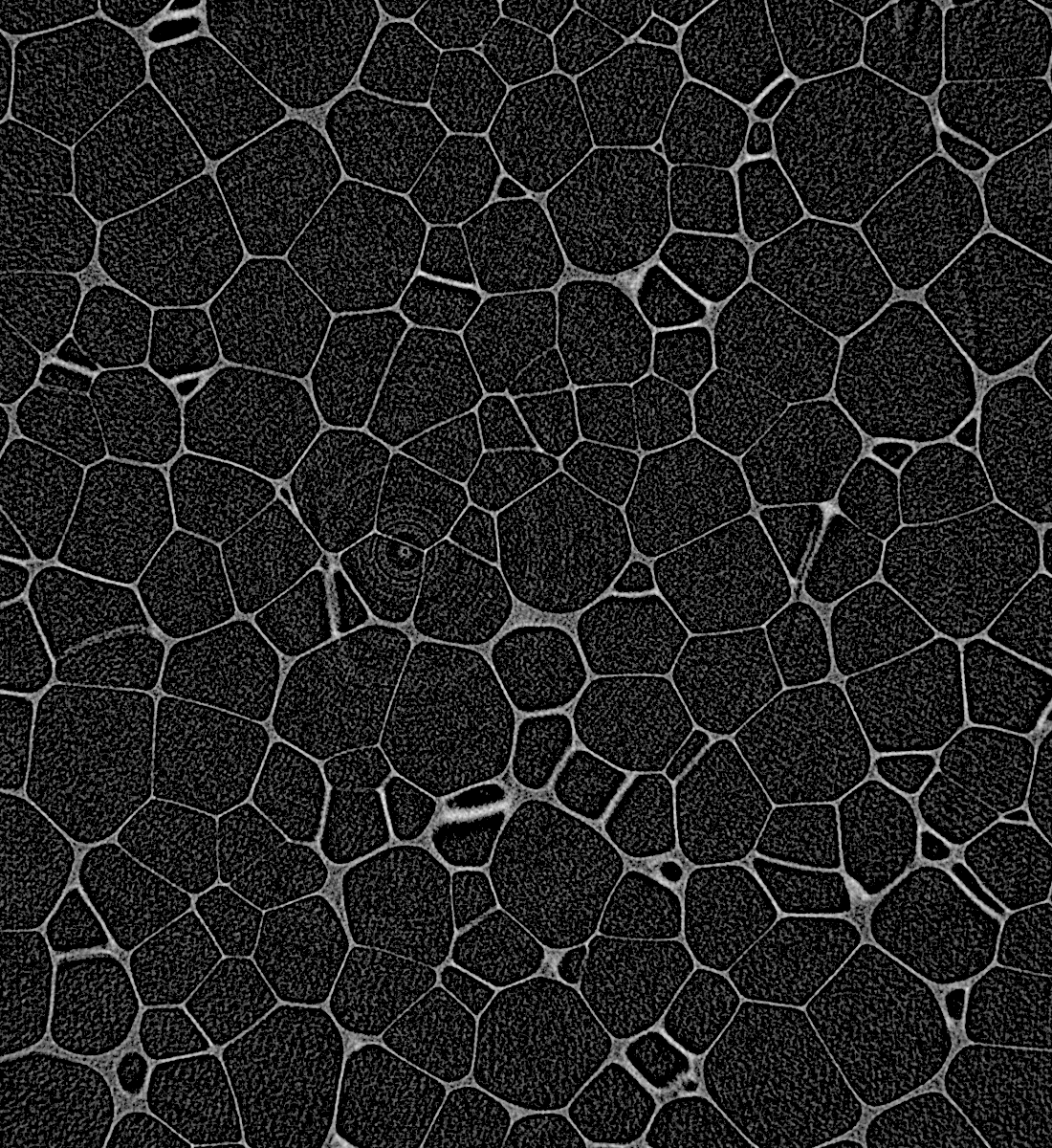}
   \includegraphics[height=0.25\textwidth]{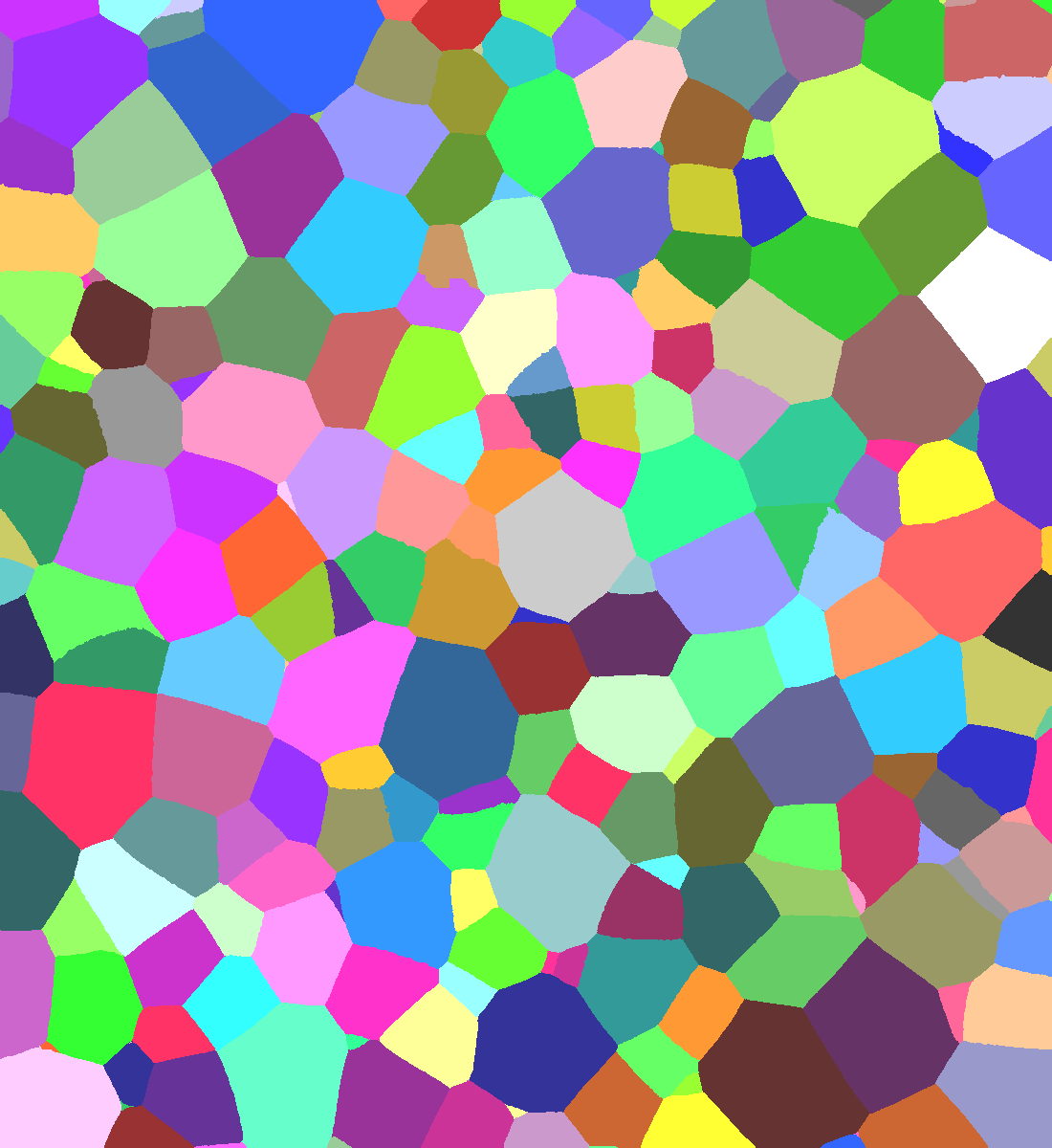}
    \includegraphics[height=0.25\textwidth]{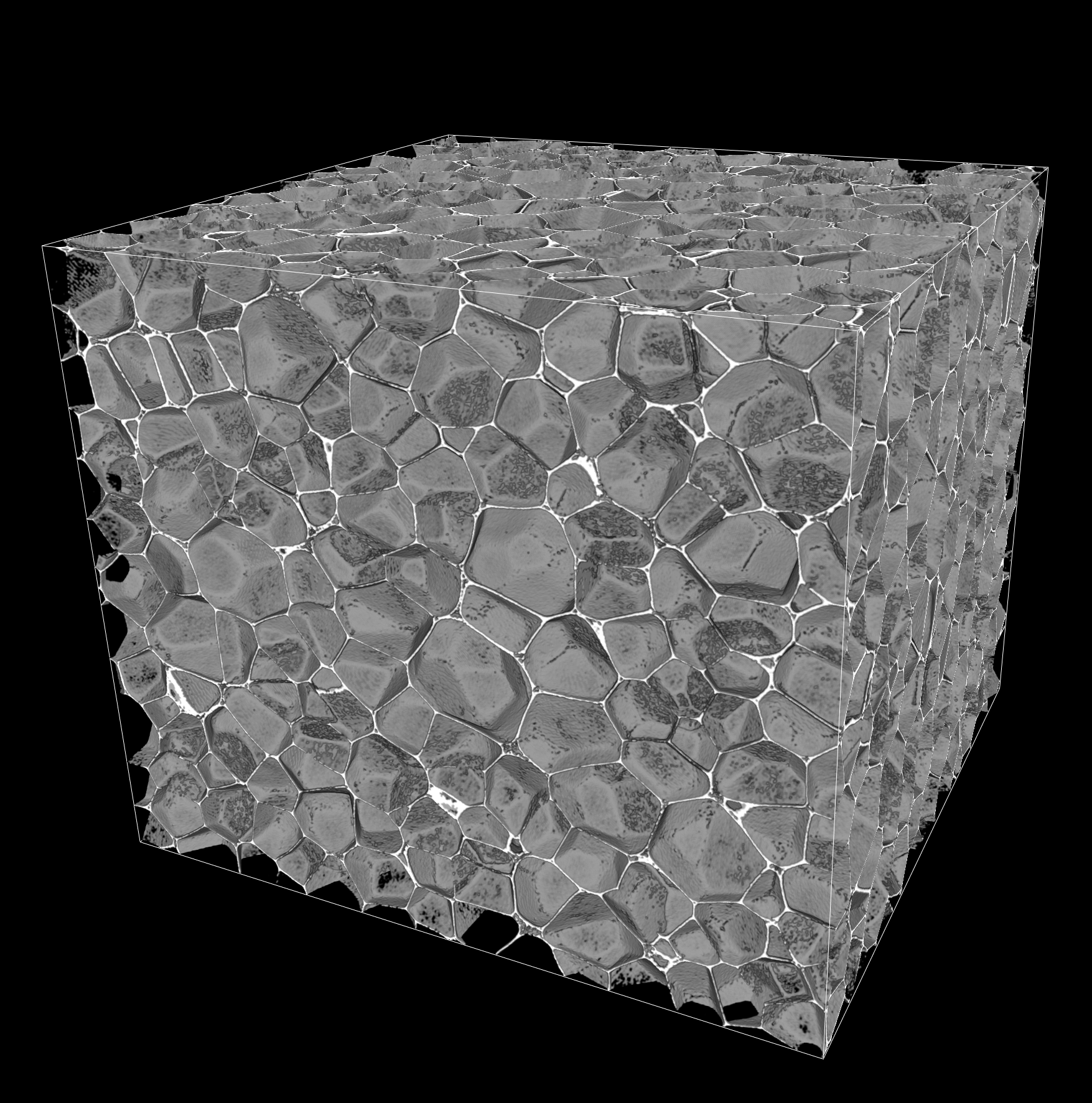}
       \caption{2D slice of the reconstructed CT image of a closed polymer foam (left), cell reconstruction by watershed (middle), and a volume rendering (right). The image size is $1300\times 1100\times 1000$ voxels with a voxel edge length of \SI{2.72}{\micro\metre}.\label{fig:PolymerFoam}}
 \end{figure}

The second example is a Rohacell\textsuperscript{\textregistered} polymethacrylimide (PMI) closed-cell foam (WIND-F RC100) that was imaged by micro computed tomography with a voxel edge length of~$\SI{2.72}{\micro\meter}.$  The image size is $1300 \times 1100 \times 1000$ voxels. Analysis of the image and Laguerre tessellation-based models for the 3D microstructure are presented in \cite{VECCHIO2014171,VECCHIO201660}. A 2D slice of the polymer foam image and its segmentation are shown in Figure \ref{fig:PolymerFoam}. Here, we consider a cropped section of the image of size $600 \times 600\times 600$ voxels that contains 380 pores. We will refer to the data set as \textbf{WIND}.

\vspace{-0.4cm}

\subsubsection{Closed zinc foam}

\begin{figure}[h]
    \centering
    \includegraphics[height=0.24\textwidth]{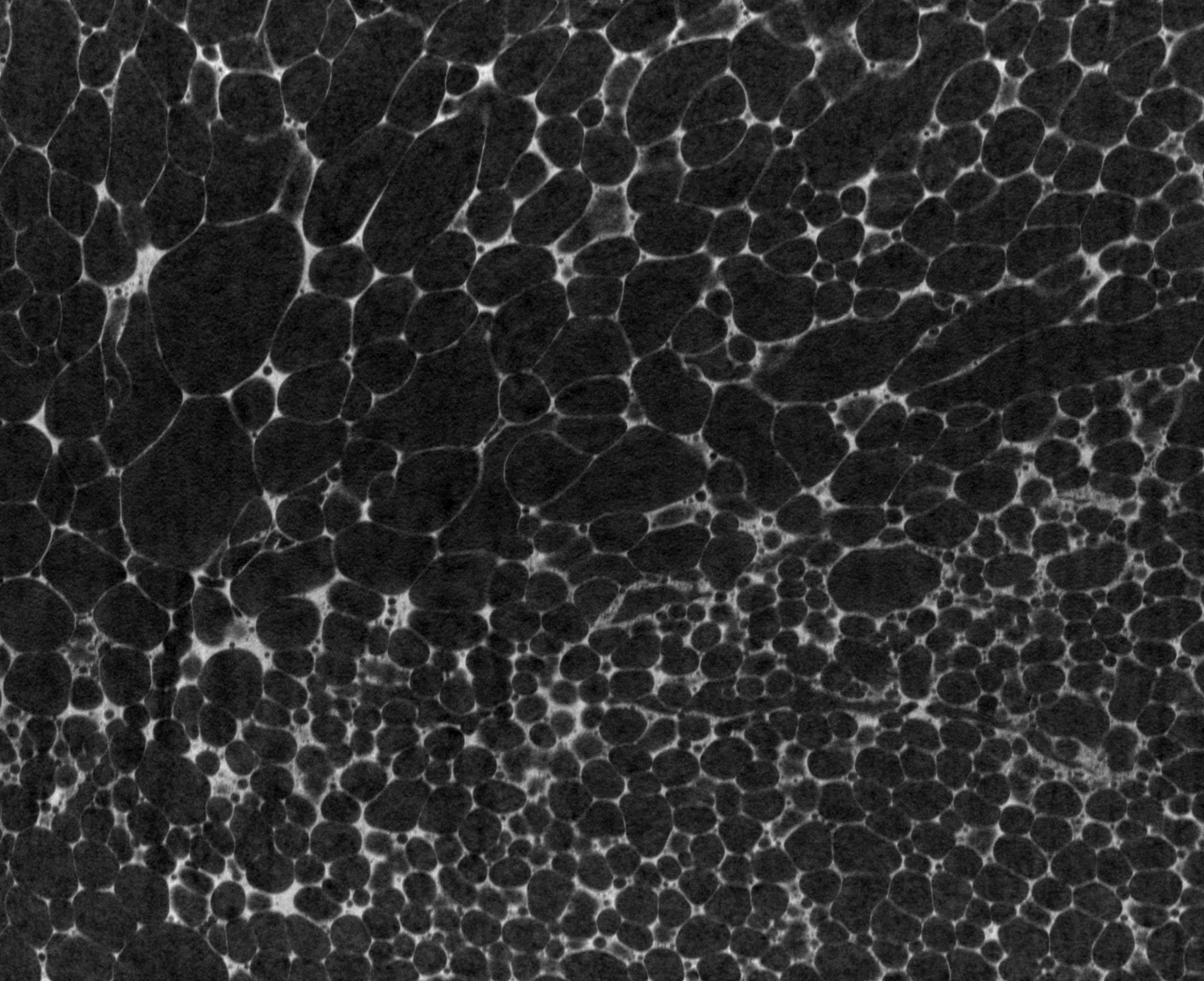}
    \includegraphics[height=0.24\textwidth]{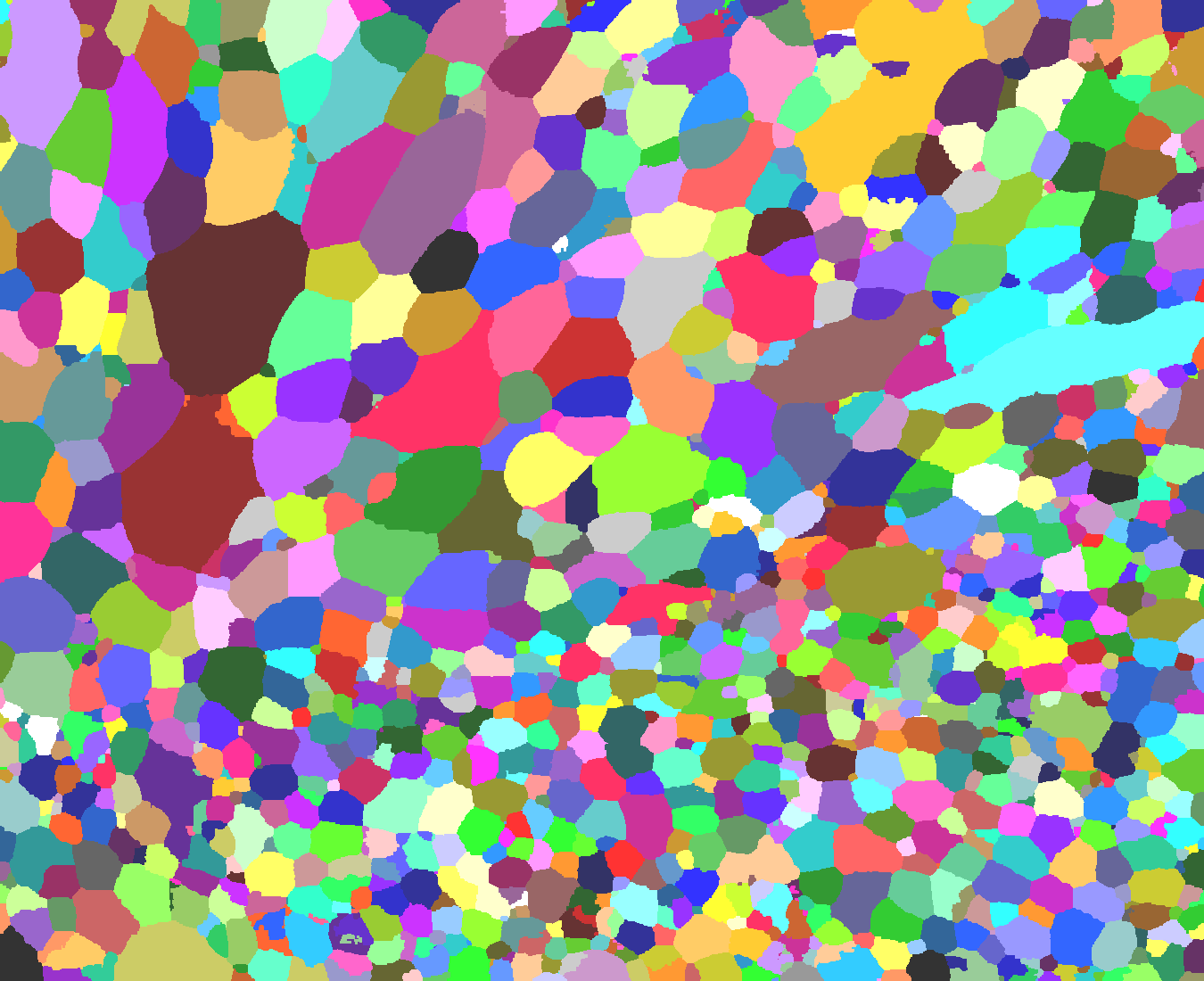}
    \includegraphics[height=0.24\textwidth]{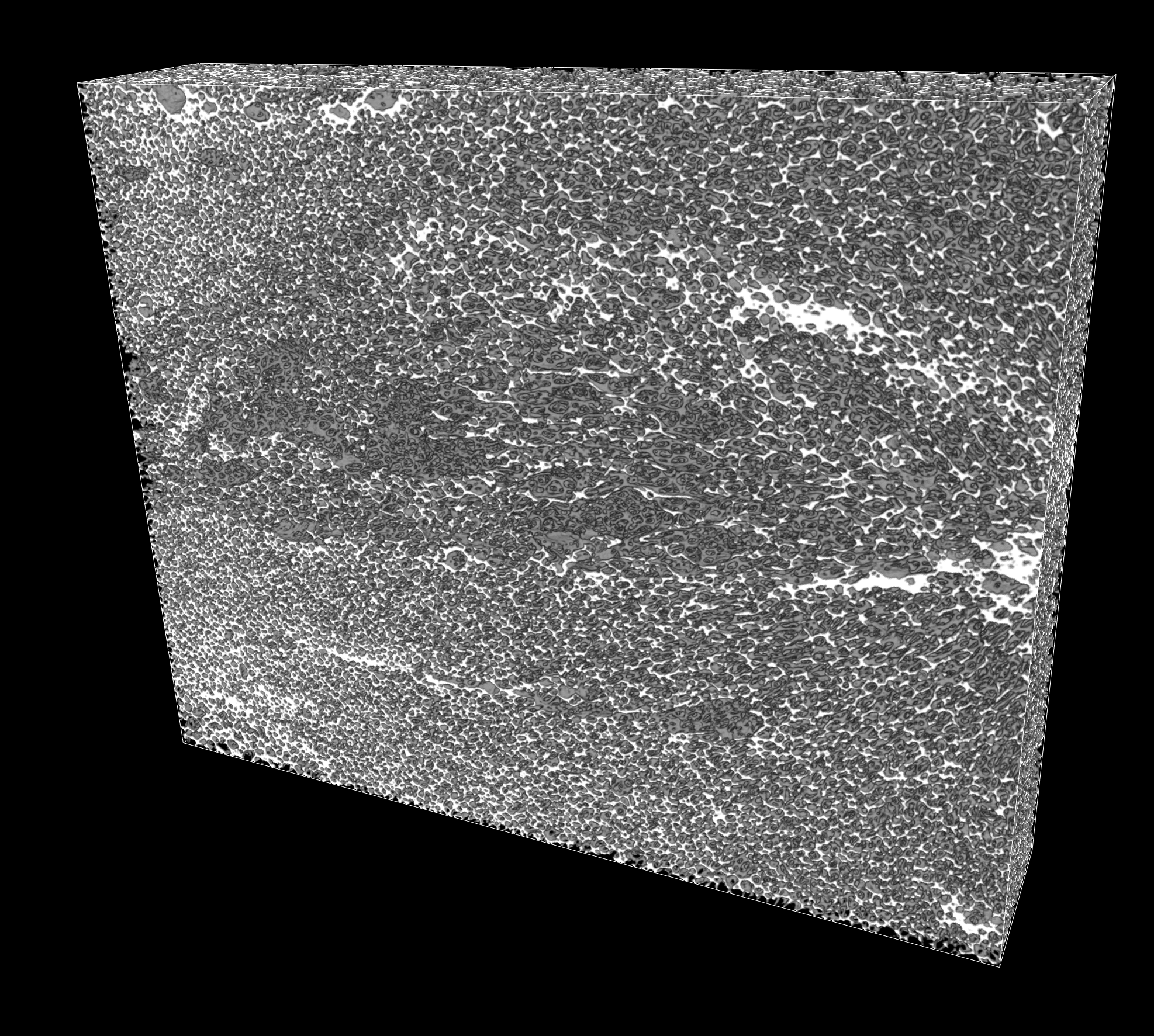}
    \caption{2D slice of the reconstructed CT image of a closed zinc foam (left), cell reconstruction by watershed (middle), and a volume rendering (right). The image size is $975\times 1100\times 1350$ voxels with a voxel edge length of~\SI{18.95}{\micro\metre}.}\label{fig:grillo}
\end{figure}

\begin{figure}[h]
    \centering
    \includegraphics[width=0.22\textwidth]{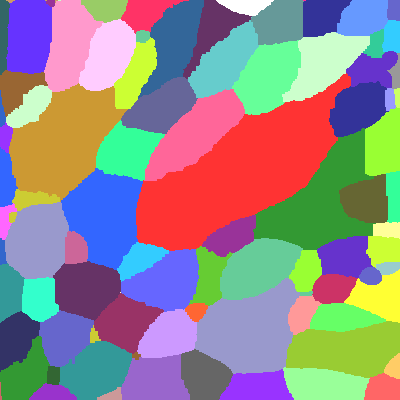}
    \includegraphics[width=0.22\textwidth]{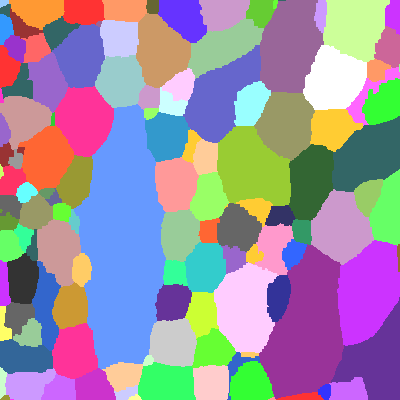}
    \caption{Cropped cutouts of the cell reconstruction shown in Figure \ref{fig:grillo} (middle), both of size $400\times 400\times 400$ voxels. Left: Zinc1, right: Zinc2}\label{fig:grillo-crop}
\end{figure}

The closed zinc foam sample and its 
image are sourced from the project \emph{Genormte Charakterisierung zellularer Werkstoffe
mittels Computertomografie }(NORMZELL, BMWi funding reference 01FS11003). The image size is $975 \times 1100 \times 1350$ voxels with a voxel edge length of~\SI{18.95}{\micro\meter}. A slice of the zinc foam image along with its segmentation are shown in Figure \ref{fig:grillo}. Since this 3D image contains significantly more pores, i.e., discrete cells, than the other data sets considered in this paper, using the full image would substantially increase the computational effort for the fitting algorithms.
Moreover, the complete 3D image is rather inhomogeneous: larger and more elongated cells occur in some regions, whereas smaller and less elongated cells are observable in other regions.
 To avoid this issue, here, we consider two cropped sections of the image, each of size $ 400\times 400\times 400$ voxels, 
 chosen to represent regions with different characteristic cell sizes. These cutouts contain 879 and 1,903 pores, respectively. The cropped cutouts are shown in Figure \ref{fig:grillo-crop}. We will refer to the data set as \textbf{Zinc1}
 and~\textbf{Zinc2}.

\vspace{-0.4cm}

\section{Comparison of fitting results}\label{sec:comparison}

\subsection{Measures for performance evaluation}\label{sect:eval}
To assess the goodness of fit between the discretized tessellation sets $C_1=C_1(\phi)\cap W,\dots,C_n=C_n(\phi)\cap W$ and the ground truth $\GT: W \to \{0,\dots,n\}$, representing $n$ grains/cells, we compute a range of performance measures.  First, we consider the fraction $F_\mathrm{c}$ of correctly assigned voxels given by
\[
	F_\mathrm{c}= \frac{1}{
	|\{x\in W \colon \GT(x)>0\}} \sum_{i=1}^n|C_i^\GT\cap C_i|.
\]
Furthermore, we consider the relative frequency $
F_0 = n^{-1}|\{i\in\{1,\dots,n\}\colon C_i=\emptyset\}|$ of missing grains/cells. 
The values for $F_\mathrm{c}$ and $F_0$ for the fitting algorithms considered are listed in Tables~\ref{table:acc} and \ref{table:missing}, respectively.

\begin{table}[h]
	\centering
	\small
\begin{tabular}{lllllllllll}
	\toprule
	& \multicolumn{1}{c}{\textbf{Voronoi}} & \multicolumn{4}{c}{\textbf{Laguerre}} & \multicolumn{1}{c}{\textbf{d-GBPD}} & \multicolumn{4}{c}{\textbf{GBPD}} \\
	\cmidrule(lr){2-2} \cmidrule(lr){3-6} \cmidrule(lr){7-7} \cmidrule(lr){8-11}
	& \textbf{GD} & \textbf{Hq} & \textbf{Neper} & \textbf{CE} & \textbf{GD} & \textbf{GD} & \textbf{GD} & \textbf{LP} & \textbf{H$_0$} & \textbf{Hq} \\ 
	\midrule
\textbf{AlCuStep0} & 0.80 & 0.88 & 0.93 & 0.86 & 0.85 & 0.93 & \textbf{0.95} & 0.90 & 0.93 & 0.93 \\
\textbf{AlCuStep20} & 0.82 & 0.89 & 0.94 & 0.88 & 0.86 & 0.94 & \textbf{0.95} & 0.93 & 0.94 & 0.94 \\
\textbf{AlCuStep60} & 0.82 & 0.89 & 0.93 & 0.90 & 0.85 & 0.94 & \textbf{0.96} & 0.93 & 0.94 & 0.94 \\
\textbf{PureFeStep0} & 0.68 & 0.69 & 0.73 & 0.62 & 0.71 & 0.80 & \textbf{0.90} & 0.85 & 0.85 & 0.86 \\
\textbf{Ceramic} & 0.91 & 0.89 & 0.92 & 0.91 & 0.91 & \textbf{0.93} & 0.90 & \textbf{0.93} & 0.92 & 0.92 \\
\textbf{WIND} & 0.90 & 0.88 & 0.94 & 0.93 & 0.92 & \textbf{0.96} & 0.94 & 0.93 & 0.93 & 0.93 \\
\textbf{Zinc1} & 0.74 & 0.74 & 0.80 & 0.74 & 0.77 & 0.86 & \textbf{0.92} & 0.90 & 0.89 & 0.90 \\
\textbf{Zinc2} & 0.76 & 0.80 & 0.83 & 0.79 & 0.80 & 0.89 & \textbf{0.91} & \textbf{0.91} & 0.90 & 0.90 \\
\bottomrule
\end{tabular}

	\caption{Values of $F_{c}$ achieved by deploying the considered fitting methods to the different data sets.}
	\label{table:acc}
\end{table}

\begin{table}[h]
	\centering
	\small
\begin{tabular}{lllllllllll}
	\toprule
	& \multicolumn{1}{c}{\textbf{Voronoi}} & \multicolumn{4}{c}{\textbf{Laguerre}} & \multicolumn{1}{c}{\textbf{d-GBPD}} & \multicolumn{4}{c}{\textbf{GBPD}} \\
	\cmidrule(lr){2-2} \cmidrule(lr){3-6} \cmidrule(lr){7-7} \cmidrule(lr){8-11}
	& \textbf{GD} & \textbf{Hq} & \textbf{Neper} & \textbf{CE} & \textbf{GD} & \textbf{GD} & \textbf{GD} & \textbf{LP} & \textbf{H$_0$} & \textbf{Hq} \\ 
	\midrule
\textbf{AlCuStep0} & 0.01 & \textbf{0.00} & \textbf{0.00} & \textbf{0.00} & 0.02 & 0.09 & \textbf{0.00} & 0.12 & \textbf{0.00} & \textbf{0.00} \\
\textbf{AlCuStep20} & \textbf{0.00} & \textbf{0.00} & \textbf{0.00} & \textbf{0.00} & 0.01 & 0.10 & \textbf{0.00} & \textbf{0.00} & \textbf{0.00} & \textbf{0.00} \\
\textbf{AlCuStep60} & \textbf{0.00} & \textbf{0.00} & \textbf{0.00} & \textbf{0.00} & \textbf{0.00} & 0.11 & \textbf{0.00} & 0.05 & \textbf{0.00} & \textbf{0.00} \\
\textbf{PureFeStep0} & 0.01 & 0.02 & 0.04 & 0.03 & 0.02 & 0.08 & \textbf{0.00} & \textbf{0.00} & \textbf{0.00} & \textbf{0.00} \\
\textbf{Ceramic} & 0.01 & \textbf{0.00} & \textbf{0.00} & \textbf{0.00} & 0.01 & \textbf{0.00} & 0.01 & \textbf{0.00} & \textbf{0.00} & \textbf{0.00} \\
\textbf{WIND} & 0.04 & 0.01 & \textbf{0.00} & \textbf{0.00} & 0.07 & \textbf{0.00} & 0.02 & \textbf{0.00} & \textbf{0.00} & \textbf{0.00} \\
\textbf{Zinc1} & 0.01 & 0.01 & 0.02 & 0.02 & 0.02 & 0.19 & \textbf{0.00} & 0.07 & \textbf{0.00} & \textbf{0.00} \\
\textbf{Zinc2} & 0.01 & \textbf{0.00} & 0.01 & \textbf{0.00} & 0.01 & 0.12 & \textbf{0.00} & 0.05 & \textbf{0.00} & \textbf{0.00} \\
\bottomrule
\end{tabular}

	\caption{Values of $F_{0}$ achieved by deploying the considered fitting methods to the different data sets.}
	\label{table:missing}
\end{table}

In addition, we compute performance measures that quantify grain/cell-wise discrepancies.  From here on, we will restrict the comparison to non-missing grains/cells. Therefore, we define the index set 
$
\mathscr{I}= \{i\in\{1,\dots,n\} \colon  C_i \neq \emptyset \}$.
To characterize the shape and size of a set $C \subseteq W$, we consider the volume-equivalent diameter $\varphi_\mathrm{d}(C)$ of $C$, which is given by
\[
\varphi_\mathrm{d}(C)= \sqrt[3]{\frac{6 |C|}{\pi}},
\] 
as well as the surface area $\varphi_\mathrm{A}(C)$, the elongation factor $\varphi_\mathrm{elo}(C)$ (ratio of second longest to longest half axes lengths of best fitting ellipsoid), and the flatness factor $\varphi_\mathrm{flat}(C)$ (ratio of shortest to second longest half axes lengths of best fitting ellipsoid) \cite{furat.2024multi}. Here, by best-fitting ellipsoid we refer to a moment-equivalent ellipsoid, i.e.,  an ellipsoid whose second-order central moments coincide with those of the corresponding cell. For details on the computation of second moments of discretized cells by means of principal component analysis, see \cite{Hastie2029}. 
Then, for each descriptor $\varphi\in\{\varphi_\mathrm{d},\varphi_\mathrm{A},\varphi_\mathrm{elo},\varphi_\mathrm{flat}\}$ we quantify the mean normalized error of the fit by
\[
	F_\varphi = \frac{1}{|\mathscr{I}| \overline{\varphi}} \sum_{i \in \mathscr{I}}
	|\varphi(C_i^{\GT})- \varphi(C_i)|,
\]
where $\overline{\varphi}= \frac{1}{n}\sum_{i=1}^{n} \varphi(G_i^{\GT})$. The values of the performance measures 
$F_{\varphi_\mathrm{d}}$, $F_{\varphi_\mathrm{A}}$, $F_{\varphi_\mathrm{elo}}$, $F_{\varphi_\mathrm{flat}}$ for the considered fitting algorithms are given in Tables~S1, S2, S3, and S4 of the Supporting Information (SI), respectively.

Furthermore, we define a performance measure to assess the accuracy with which the sets $C_1,\dots,C_n$ of the fitted tessellation reflect the topology of GT.
 Therefore, for each $i\in\mathscr{I}$, let $N^\GT(i), N^T(i) \subseteq \{1, \dots, n\}$  denote the sets of indices of grains/cells adjacent to $C_i^{\GT}$ and $C_i$ within the ground truth GT and the fitted tessellation, respectively, where the notion of adjacency is based on the $ \mathcal{N}_{26} $ neighborhood, as in Section~\ref{sect:CE}.  We then quantify the agreement between the neighborhood of $C_i^{\GT}$ and that of $C_i$ using the intersection over union $\mathrm{IoU}(i)$ defined as 
\[
	\mathrm{IoU}(i)= \frac{|N^\GT(i) \cap N^T(i)|}{|N^\GT(i) \cup N^T(i)|}.
\]
Finally, we quantify the goodness-of-fit with respect to the topology by the performance measure $F_\mathrm{IoU}$ which is given by
\[
	F_\mathrm{IoU}= \frac{1}{|\mathscr{I}|} \sum_{i \in \mathscr{I}} \mathrm{IoU}(i).
\]
The values of $F_\mathrm{IoU}$ for the fitting algorithms considered are listed in Table~S5 of the SI.

\subsection{Discussion}\label{sect:results}
The values for the volume-based accuracy measure $F_\mathrm{c}$ listed in Table~\ref{table:acc} indicate that many differences between the fitting algorithms considered in the present paper could be attributed to the choice of the tessellation model, rather than the fitting method. 
Overall, methods that fit GBPD-type tessellations generally achieve higher accuracies than those that fit Voronoi and Laguerre tessellations.   This holds particularly for datasets that exhibit more complex grain/cell morphologies (e.g., \textbf{Zinc1}, \textbf{PureFeStep0}). This is mostly to be expected because of the higher flexibility of GBPDs. However, for datasets that exhibit `simpler' grain/cell morphologies (e.g., convex grains/cells), such as the \textbf{Ceramic} and \textbf{WIND} datasets, the algorithms that fit Laguerre tessellations already perform quite well. This could be attributed to the fact that Laguerre tessellations inherently enforce the constraint of planar grain/cell facets. Among the algorithms that fit Laguerre tessellations, Neper yields the best accuracy values for the considered datasets. Notably, for the two datasets \textbf{Ceramic} and \textbf{WIND}, the fit achieved by the even simpler Voronoi tessellation yields results comparable to those of Laguerre tessellations fitted with Neper (see Table~\ref{table:acc}). As expected, for datasets that exhibit more complex grain/cell architectures, the more general Laguerre tessellation outperforms the Voronoi tessellation due to its greater flexibility in controlling grain/cell sizes through an additional parameter per grain/cell.

In total, GBPD tessellations fitted by the GD method, which fits all parameters, yield the best performance with respect to $F_\mathrm{c}$ for most of our datasets. In fact,  across our datasets, the accuracy of the evaluated GBPD algorithms varies by no more than 5 percentage points.  Interestingly, the Hq algorithm achieves the lowest mean normalized errors \( F_{\varphi_d} \) in volume-equivalent diameters across the evaluated data sets. This suggests that its underlying heuristic tends to generate tessellations whose volumes of the sets $C_i(\phi)$ closely match those observed in the data. A similar trend was observed in~\cite{alpers.2023a}, although that study also identified grain/cell configurations for which Hq does not produce satisfactory fits. In contrast, the LP method ensures that the generated GBPDs adhere to specified volume bounds, which were set to $\pm2$ voxels per grain/cell in our experiments. When these bounds are reduced to zero, LP---unlike the other methods listed in Table~S1 of the SI---yields a volume-equivalent diameter error of $F_{\varphi_d} = 0$ by construction.

The surface area errors $F_{\varphi_A}$ in Table~S2 of the SI show a similar trend as in Table~\ref{table:acc}. In particular, Laguerre tessellations have a limited capacity to reproduce the curved facets of the more complex data sets. Again, Laguerre tessellations fitted by Neper show a good (even the best) performance for the two foam data sets \textbf{Ceramic} and \textbf{WIND}.

The better ability of GBPDs to fit cell shapes is also evident from the errors $F_{\varphi_\mathrm{elo}}$ and $F_{\varphi_\mathrm{flat}}$ of the shape characteristics elongation and flatness, see Tables~S3 and S4 of the SI.

For almost all the considered datasets, Hq has the smallest frequency of missing grains/cells, see Table~\ref{table:missing}. This can be explained by the fact that systems of non- or only mildly overlapping objects do not generate empty sets. Hence, the construction used in the heuristics---sites corresponding to the barycenters of the grains/cells---is not prone to produce sets of generators with empty sets $C_i(\phi)$. The choice of update scheme in the iterative algorithms determines to which extent empty sets $C_i(\phi)$ can arise in subsequent iterations.

\begin{figure}[h]
    \centering
    \begin{tabular}{ccc}
        \includegraphics[width = 0.3\textwidth]{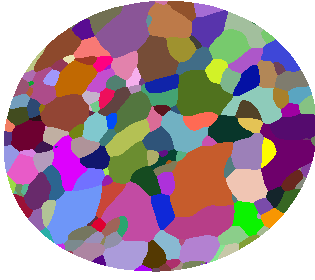} & 
        \includegraphics[width = 0.3 \textwidth]{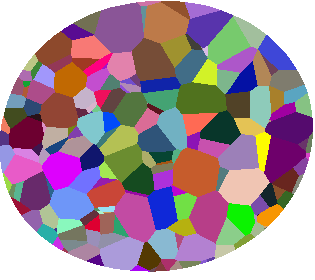} 
        & 
        \includegraphics[width = 0.3\textwidth]{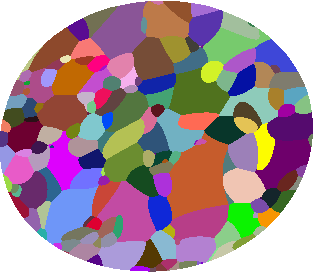} 
    \end{tabular}
    \caption{Fitting of the data set described in Section~\ref{sec:DataAA} using the method described in Section~\ref{method:gradient-descentI}. Results are visualized by comparing image data~(left), the fit with a Laguerre tessellation~(center), and the fit of a GPBD~(right) at the example of one 2D slice.} 
    \label{fig:enter-label}
\end{figure}

With respect to topology, GBPDs outperform Laguerre tessellations again, see Table~S5 of the SI. The latter cannot reproduce neighborhood relations that require curved boundaries, see Figure~\ref{fig:enter-label}.

\vspace{-0.2cm}

Cross sections of the  data sets considered in this paper, together with a fitted Voronoi tessellation using GD,
fitted Laguerre tessellations using Hq, Neper, CE and GD, a fitted diagonal
GBPD using GD, and fitted GBPDs using GD, LP, H$_0$ and Hq, are shown in Figures~S1 to S8 of the SI. A rough estimate of hardware-specific runtimes can be found, for example, in \cite{petrich.2021}.

\vspace{-0.2cm}

\section{Summary}\label{sect:summary}
We reviewed various algorithms for fitting Voronoi tessellations, Laguerre tessellations and GBPDs to grain and cell structures observed in  polycrystalline materials and foams. The effectiveness of these algorithms varied depending on the specific data sets, the tessellation models employed, and the evaluation criteria used. 
The heuristic Hq achieved good performance on our data sets while incurring small  computational cost. However, this is not always the case—as illustrated in~\cite{buze-pyAPD}, where the relative error in the resulting areas reaches 380\%, and in~\cite{alpers.2023a}, which highlights substantial differences between Hq and~H$_0$. Nevertheless, due to its very low computational cost, Hq seems a useful choice for generating initial configurations for further optimization.  
Each of the presented algorithms exhibits distinct characteristics, and in practical applications, one must determine which of these are most suitable for the specific use case. The modular structure of the proposed fitting approaches enables targeted improvements, such as enhancing the accuracy of surface area or volume representations. The established discrepancy metrics primarily account for mismatches in grain/cell volumes and boundaries. In addition, metrics related to grain/cell shape and topology—used here for model validation—can also be incorporated directly into the fitting process. This flexibility allows for the development of application-specific fitting strategies tailored to the structural features of interest.

\section*{Acknowledgements}
The authors thank Carl E. Krill III and Henning F. Poulsen for providing tomographic image data.
CJ and CR acknowledge funding provided by the Research Initiative of the Federal State of Rhineland-Palatinate (Potential Area MSO). AA and AS acknowledge funding from the Engineering and Physical Sciences Research Council (EPSRC grant no. EP/X035883/1). OF, MN and VS acknowledge financial support from the German Research Foundation (DFG, 673729),
and  the German Federal Ministry of Research, Technology and Space (BMFTR, 01IS21091) within the French-German research project SMILE. Moreover, the contribution of MN and OF is based on work from COST
Action 24122 mSPACE, supported by COST (European Cooperation in Science
and Technology), www.cost.eu. This work has benefitted from Dagstuhl Seminar 25492 "Generalized Voronoi Diagrams and Applications."

\bibliography{Literature}
\bibliographystyle{unsrt}

\end{document}